\pdfoutput=1

\documentclass[11pt]{article}

\usepackage[]{ACL2023}

\usepackage{times}
\usepackage{latexsym}

\usepackage[T1]{fontenc}

\usepackage[utf8]{inputenc}

\usepackage{microtype}

\usepackage{graphicx}
\usepackage{comment}
\usepackage{algorithm,algcompatible,amsmath}
\usepackage{algpseudocode}

\usepackage{inconsolata}

\usepackage{natbib}
\usepackage[english]{babel}
\usepackage{blindtext}
\usepackage{tabularx}
\usepackage{bold-extra}
\usepackage{graphicx}

\usepackage{array,booktabs,ragged2e}
\usepackage{longtable}

\title{RAGAR, Your Falsehood Radar: \\RAG-Augmented Reasoning for Political Fact-Checking using \\ Multimodal Large Language Models}

\author{Mohammed Abdul Khaliq\textsuperscript{1,3}, Paul Yu-Chun Chang\textsuperscript{2}\footnotemark[1], \\
\bf{Mingyang Ma\textsuperscript{2}}, \bf{Bernhard Pflugfelder\textsuperscript{2}}, \bf{Filip Miletić\textsuperscript{1}\footnotemark[1]} \\
  \textsuperscript{1}Institute for Natural Language Processing, University of Stuttgart, \\
  \textsuperscript{2}appliedAI Initiative GmbH, \textsuperscript{3}appliedAI Institute for Europe gGmbH\\
  \texttt{\{mohammed.abdul-khaliq, filip.miletic\}@ims.uni-stuttgart.de},\\
  \texttt{\{p.chang, m.ma, b.pflugfelder\}@appliedai.de}\\}

\makeatletter
\def\thanks#1{\protected@xdef\@thanks{\@thanks
        \protect\footnotetext{#1}}}
\makeatother

\begin{document}
\maketitle

\begingroup
\renewcommand{\thefootnote}{\fnsymbol{footnote}}
\footnotetext[1]{Corresponding Author}
\endgroup

\begin{abstract}
The escalating challenge of misinformation, particularly in political discourse, requires advanced fact-checking solutions; this is even clearer in the more complex scenario of multimodal claims. We tackle this issue using a multimodal large language model in conjunction with retrieval-augmented generation (RAG), and
introduce two novel reasoning techniques: Chain of RAG (CoRAG) and Tree of RAG (ToRAG).
They fact-check multimodal claims by extracting both textual and image content, retrieving external information, and reasoning subsequent questions to be answered based on prior evidence.
We achieve a weighted F1-score of 0.85, surpassing a baseline reasoning technique by 0.14 points.
Human evaluation confirms that the vast majority of our generated fact-check explanations contain all information from gold standard data.
\end{abstract}

\section{Introduction}
In the age of digital information, rapid dissemination of news, both genuine and fabricated, has become a defining feature of public discourse. The phenomenon of fake news -- which more precisely denotes misinformation, disinformation, or a combination of both \citep{aimeur_fake_2023} -- 
is particularly prevalent on social media: false information spreads six times faster than the truth on platforms like Twitter \citep{vosoughi2018}. This trend poses a critical challenge to the democratic process since it makes voters increasingly prone to making decisions based on incorrect information.
The matter is further aggravated by visual information, which provides yet another widespread and consequential source of fake news. For instance, fake news stories that include images spread further than those containing only text \cite{zannettou2018origins}.


    \begin{figure}[ht]
    \centering
    \includegraphics[width=7.5cm]{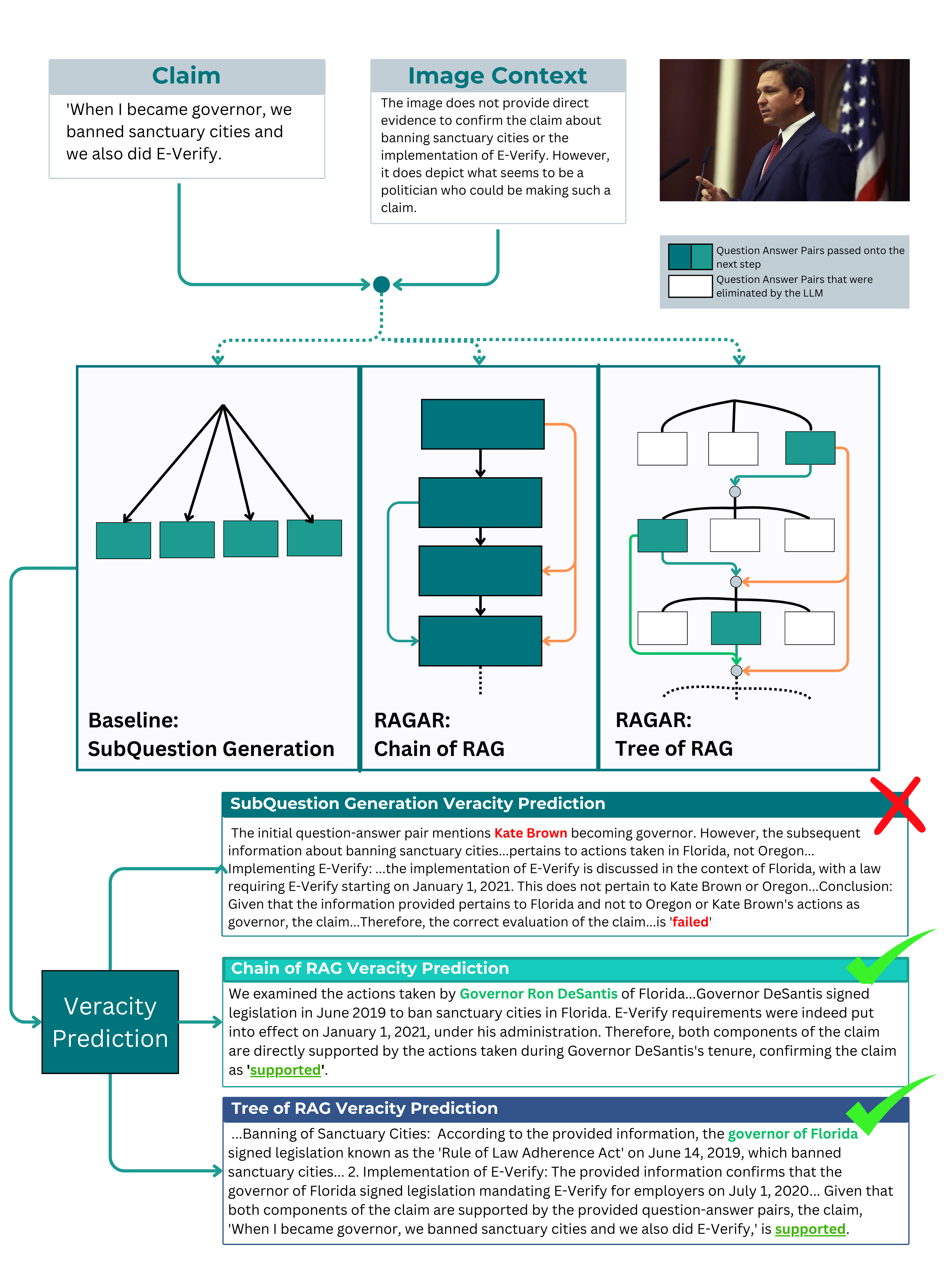}
    \caption{An overview of the fact-checking pipeline contrasting the baseline Sub-Question Generation approach from the Chain of RAG and Tree of RAG approach followed by veracity prediction and explanation.}
    \label{fig:highlevelpipeline}
\end{figure}

A potential solution to these issues is provided by automated fact-checking systems.
They have benefited from the development of large language models (LLMs),
leading to improvements in detection, labeling, and generation of veracity explanations \citep{DAS2023103219}.
More recently, multimodal approaches 
have complemented textual information with image representations to assess their cross-modal consistency and unified embedding representations \citep{Yao_2023}. 
Another active line of research deploys retrieval-augmented generation (RAG), whereby LLMs access up-to-date external information at inference time.
They convert the input claim into phrase queries, pass them onto a search engine, and use the retrieved information to assess veracity \citep{asai2024selfrag, Zeng2024JustiLMFJ}.
It however remains to be determined if more elaborate reasoning techniques can be beneficial in this setting.
Moreover, RAG-based approaches have so far mostly been applied to text.
This raises the additional question of their use in the more challenging scenario of multimodal fact-checking.

Addressing this gap, we introduce RAGAR -- RAG-Augmented Reasoning techniques, which we apply to multimodal fact-checking in the political domain (see Figure~\ref{fig:highlevelpipeline} for a high-level overview).
We rely on a multimodal LLM to verbalize the textual and visual elements of a claim, and use RAG responses to motivate successive steps in determining veracity.
The system is underpinned by elaborate reasoning strategies instantiated in two distinct approaches: Chain of RAG (CoRAG) and Tree of RAG (ToRAG).
We evaluate them using a multimodal fact-checking dataset as well as human annotation of generated explanations.

Our contributions are as follows.
(1)~We introduce two novel reasoning techniques for multimodal fact-checking, reaching a weighted F1-score of 0.85.
(2)~We provide two complementary strategies for multimodal input: during claim generation, we verbalize image content with respect to the associated text; during retrieval, we look up image captions and use them as further evidence.
(3)~We run a multi-rater annotation of generated fact-check explanations, showing that the vast majority of them include all information from the gold standard.
To our knowledge, this is the first study to incorporate multimodal LLMs in a RAG-based reasoning approach applied to multimodal fact-checking for the political domain.

\section{Related Work}

\subsection{Retrieval-Augmented Generation (RAG) for Fact-Checking} \label{raggenanswersrw}

To combat hallucination in text generation, current fact-checking pipelines often implement a RAG approach, wherein an LLM retrieves data from external sources to enhance its response 
and move past its knowledge cutoff. 
\citet{peng2023check} present LLM-Augmenter, which combines external knowledge integration and automated feedback mechanisms. 
\citet{chern2023factool} assess the factuality of LLM-generated text on multiple tasks and domains,
e.g.\ for Knowledge Based Question Answering they use Google Search API to extract relevant knowledge and then parse the result. 
\citet{pan-etal-2023-fact} 
rely on LLM's in-context learning, and use Chain of Thought \cite{chainofthought} reasoning to guide the model in complex tasks such as fact-checking on the web. 
\citet{zhang2023llmbased} propose Hierarchical Step-by-Step (HiSS) prompting, which splits a claim into sub-claims, creating a hierarchy, and verifies each one through multiple question-answering steps using web-retrieved evidence. 
\citet{xu2023search} propose SearChain. It creates a Chain of Query (CoQ) reasoning chain, where each question follows from the knowledge gathered in the previous question; uses information retrieval (IR) to verify the answer at each node; and prompts the LLM to indicate missing information, which is handled by an IR call. 

Our RAGAR approaches employ a more sophisticated reasoning framework with multiple rounds of sequential question-answering, elimination (in the case of ToRAG), and verification. 
We also extend domain coverage through multimodality, and propose a zero-shot (rather than few-shot) approach.

\subsection{Multimodal Fact-Checking using LLMs} \label{multimodalrw}
Multimodality is generally underexplored in fact-checking \citep{alam-etal-2022-survey}, but several recent approaches have been proposed.
%
\citet{guo2023texts} 
use LLM-agnostic models to generate textual prompts from images and then guide LLMs in generating responses to Visual Question Answering queries. 
\citet{Yao_2023} construct a multimodal dataset using fact-checking websites, and then develop a fact-checking and explanation generation pipeline.
It encodes and reranks each sentence in the document corpus in relation to the claim, and uses a CLIP \citep{radford2021learning} encoding for images; the similarity between an input claim and the provided images is then computed. An attention model is used for multimodal claim verification, and BART \citep{lewis2019bart} for explanation generation. 
In concurrent research, \citet{chainofaction} propose the Chain of Action prompt.
It splits an input query into sub-questions and uses a ‘‘Missing Flag'' indicator to fill in or correct the answers generated by internal LLM knowledge via RAG.

Our RAGAR approaches similarly use a multimodal LLM \citep[GPT-4V;][]{openai2023gpt4v} to add context to the textual claim, but employ a different set of reasoning techniques.
We furthermore introduce a multimodal RAG component during evidence retrieval, using captions of matching images to provide the LLM with relevant meta information.

\begin{figure*}[ht]
    \centering
    \includegraphics[width=10cm]{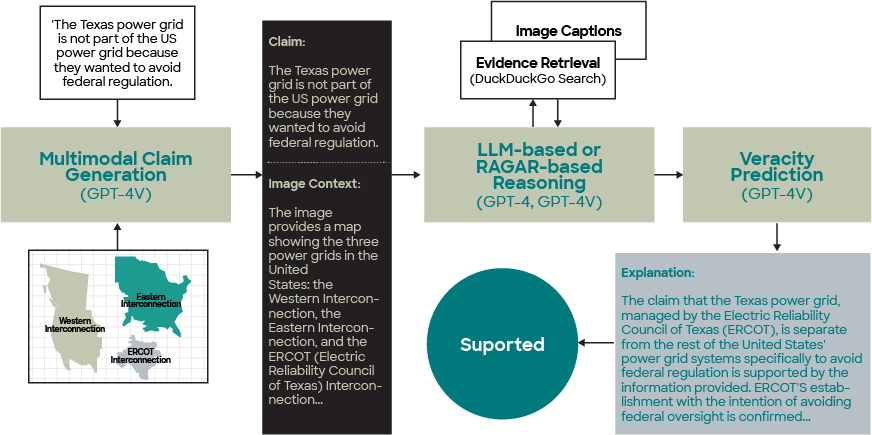}
    \caption{A detailed overview of the Multimodal Fact-checking pipeline}
    \label{fig:mmfcpipeline}
\end{figure*}

\section{Dataset}

The aim of our study is to explore the potential of multimodal LLM-based RAG and reasoning for political fact-checking.
Given the substantial computational and financial costs of running multimodal LLMs through multiple rounds of reasoning, we evaluate our approach on a well-controlled and balanced dataset, so as to minimize noise while maintaining the validity of our experiments.

We specifically rely on a carefully selected subset of the MOCHEG dataset \cite{Yao_2023}.
MOCHEG provides 21,184 multimodal claims sourced from two fact-checking websites, PolitiFact\footnote {\url{https://www.politifact.com/}} and Snopes.\footnote{\url{https://www.snopes.com/}}
Each instance contains an input claim extracted from the title of the fact-checking source, and an associated image extracted from the web page that addresses the claim. The dataset further provides a summary of the fact-check in the form of a ‘‘Ruling Outline'', which we consider for evaluating LLM-generated explanations.

We start from the test set containing 2,007 multimodal claims and filter it in two steps.
First, we select the 794 claims that were fact-checked by PolitiFact, since our focus is on political claims; 
by contrast, Snopes provides fact-checks for a variety of domains.
Second, we filter this set down to 300 test samples randomly selected from the \textit{supported} and \textit{refuted} classes, for a balanced final dataset with 150 multimodal claims in each of the two classes.

In this process, we purposefully discard the \textit{NEI} (Not Enough Information) instances.
During the creation of MOCHEG, some ambiguous cases were outright discarded, while the labels \textit{mixture}, \textit{unproven}, and \textit{undetermined} were aggregated under \textit{NEI}.
This class is potentially unstable in two respects:
fact-checking websites update their labels as new evidence emerges \cite{Yao_2023}, which by definition affects this class more prominently;
and the fact-checking intentions behind mixed labels such as \textit{half-true} and \textit{mixture} are comparatively unclear, leading prior studies to exclude them \citep[e.g.][]{vo2019learning}.
We adopt the same decision given our focus on an initial validation of novel reasoning techniques. 

Although we only retain instances that are unambiguous in the dataset, our model may still struggle to retrieve information of sufficient quality to fact-check them. We account for this
by allowing it to generate a \textit{failed} label when it fails to retrieve relevant information. 
We reserve an extension of our study to the \textit{NEI} class, as well as the connected issue of improving retrieval quality, for future work.

\section{Multimodal Fact-Checking Pipeline}
Our fact-checking pipeline comprises four parts: (i)~Multimodal Claim Generation, which analyzes both the textual claim and associated image to formulate a new claim incorporating both; (ii)~Multimodal Evidence Retrieval, which extracts evidence from the web for a question posed by the LLM; (iii)~LLM-based and RAG-augmented Reasoning for fact-checking,  our reasoning approach to fact-check a claim; and (iv)~Veracity Prediction and Explanation. The pipeline is shown in Figure~\ref{fig:mmfcpipeline}.


\subsection{Multimodal Claim Generation}
\label{multimodal-claim-generation}
Given an input claim as text, an associated image, and the date of the claim, the claim generation module generates a response verbalizing the information contained in both the textual claim and the image. We use GPT-4V as our multimodal LLM given its strong performance across tasks.
Note that our aim is not to determine the best-performing model on our task, but rather to evaluate different reasoning techniques. We therefore use the same model across experiments.

\begin{figure*}[ht]
    \centering
    \includegraphics[width=12cm]{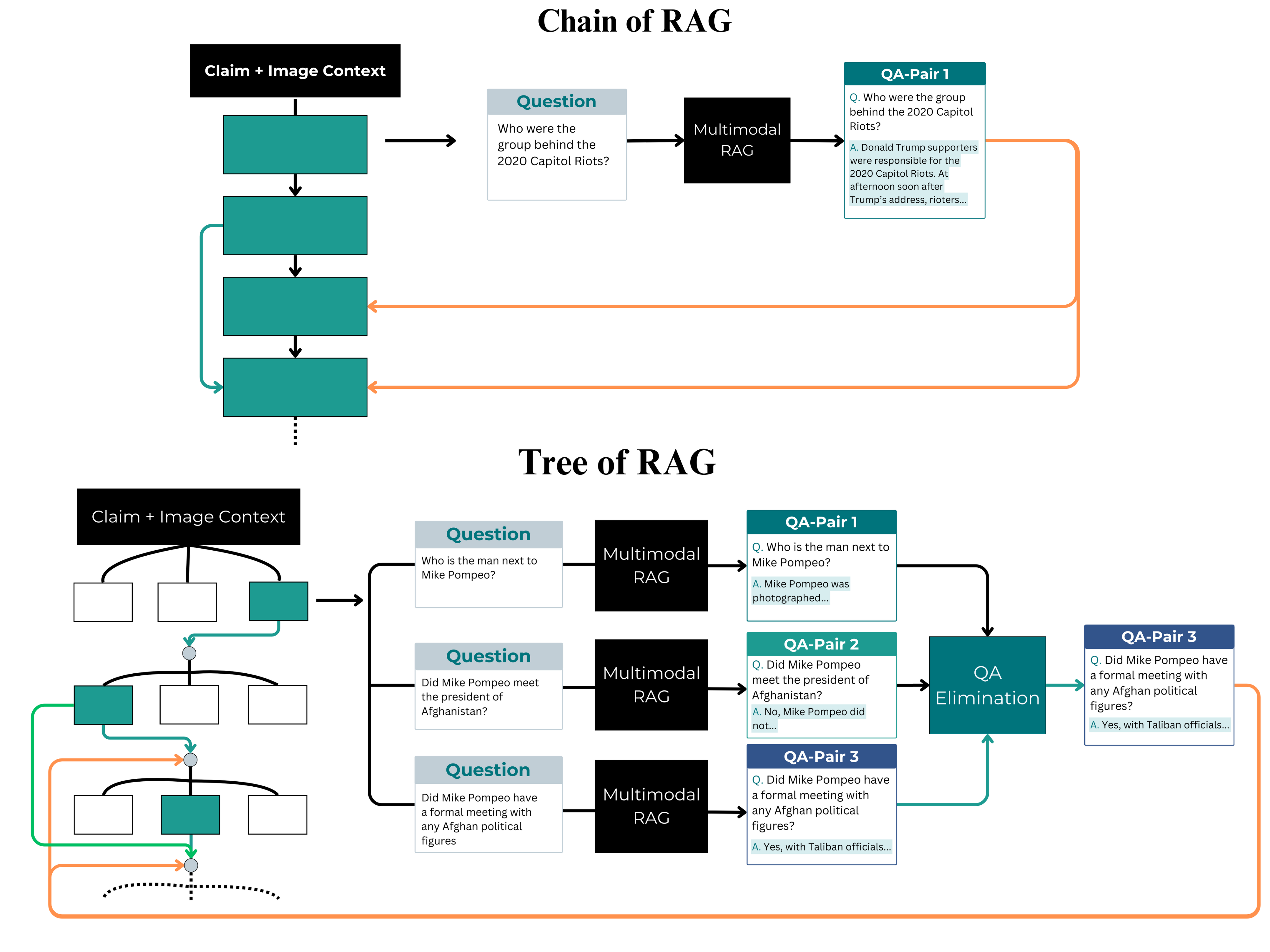}
    \caption{Chain of RAG and Tree of RAG pipeline}
    \label{fig:corgatorag}
\end{figure*}

The generated response is divided into two sections: \textit{claim}, which contains the original text claim; and \textit{image context}, which contains the details relevant to the claim extracted from the image by GPT-4V. The \textit{image context} expands on the information from the textual claim by e.g.\ identifying the speaker that the claim is quoting, extracting numerical information from figures, and highlighting relevant textual data mentioned in the image. More generally, the contextualization provides details on whether the image is relevant to the text claim. 

While directly encoding images is a potential alternative to our approach, we decide against it to allow our Chain of RAG and Tree of RAG approach to be multimodal-agnostic. This decision ensures that our reasoning methods can also be replicated with LLMs that are not inherently multimodal. 
Multimodal Claim Generation is the only section of our pipeline requiring a multimodal LLM; all remaining parts, including our RAGAR approaches, can be implemented using other LLMs and possibly extended to different tasks.
%

\subsection{Multimodal Evidence Retrieval}
\label{multimodal-evidence-retrieval}
The fact-checking questions generated by the LLM-based or RAG-augmented reasoning techniques serve as input for the multimodal evidence retrieval module. It helps answer each question by retrieving relevant text snippets from websites and further analyzing details associated with the image. 

The query to the multimodal evidence retrieval is a question generated by an LLM-based or RAGAR-based reasoning technique (presented in detail in Section~\ref{ragarexplained}). For text-based evidence retrieval, we use the DuckDuckGo Search tool provided by LangChain\footnote{\url{https://www.langchain.com/}}. We retrieve the top 10 results from the API and use them to answer the question. We temporally restrict the search by only collecting articles published in the two years before the claim was fact-checked by PolitiFact, so as to provide the LLM with facts relevant to the time-frame of the fact-check. To mimic a real-time fact-checking scenario, we remove search results that originate from \texttt{www.politifact.com}, \texttt{www.snopes.com}, and \texttt{www.factcheck.org}, since it is likely that they already contain answers to the claim and would thus impact the fairness of the experiment. We also remove the following social media websites due to potentially biased or unreliable information: \texttt{www.tiktok.com}, \texttt{www.facebook.com}, \texttt{www.twitter.com} and \texttt{www.youtube.com}. 

Most images in our dataset contain faces of politicians, pictures from political events, government buildings etc. In such cases, the image itself may not provide much additional information beyond the text claim. However, 
it is useful to determine the metadata associated with the image, which may indicate when or where the claim was made. For this purpose, 
we use SerpAPI\footnote{\url{https://serpapi.com/}} to conduct a reverse image search over the images associated with the claims. We extract the captions for the images from the first 10 results and use them as additional information for GPT-4V. This allows the model to not only analyze the image when answering an image-based question, but also incorporate meta-information about it and in that way better contextualize the answer. We demonstrate a few examples of this in Appendix \ref{multimodalragdisc}.


\subsection{LLM-Based and RAG-Augmented Reasoning for Fact-Checking}
\label{ragarexplained}

\subsubsection{Baseline: Sub-questions with Chain of Thought at Veracity Prediction (SubQ+CoT\textsubscript{VP})}
As a baseline reasoning-based approach, we employ sub-question generation followed by Chain of Thought veracity prediction (SubQ+CoT\textsubscript{VP}). This baseline is based on recent approaches to fact-checking relying on LLMs \citep{pan-etal-2023-fact, chern2023factool} as discussed in Section \ref{raggenanswersrw}. We adapt the approach to handle multimodal claims as well.

\begin{algorithm*}
\caption{Chain of RAG (CoRAG)}
\begin{algorithmic}[1]
\scriptsize
\State \textbf{Input:} Claim $C$, Image Context $I$, Image Captions $IC$
\State $Q \gets \text{GenerateFirstQuestion}(C, I)$
\State $\textbf{QAPairs} \gets \text{[]}$ \Comment{Initialize an empty list for Q-A pairs}
\State $counter \gets 0$
\State $followUpNeeded \gets \text{True}$
\While{$counter < no\_of\_steps$ \textbf{and} $followUpNeeded$}
    \If{$\text{QuestionAboutImage}(Q)$}
        \State $A \gets \text{ImageQA}(Q, I, IC)$ \Comment{Using image, question, and captions}
    \Else
        \State $A \gets \text{WebQA}(Q)$ \Comment{Standard evidence retrieval}
    \EndIf
    \State $\textbf{QAPairs}.\text{append}((Q, A))$ \Comment{Store the Q-A pair of this iteration}
    \State $followUpNeeded \gets \text{FollowupCheck}(Q, A)$
    \If{$followUpNeeded$}
        \State $Q \gets \text{FollowupQuestion}(QAPairs)$
    \EndIf
    \State $counter \gets counter + 1$
\EndWhile
\State \textbf{return} $\textbf{QAPairs}$ \Comment{Returns the list of Q-A pairs}

\end{algorithmic}
\label{alg:corag}
\end{algorithm*}

\subsubsection{RAG-Augmented Reasoning: Chain of RAG (CoRAG)}

The first novel reasoning approach we propose is Chain of RAG (CoRAG). It builds upon general RAG approaches by using sequential follow-up questions -- augmented from the RAG response -- to retrieve further evidence. 
In other words, we follow a decomposed setup, guiding the LLM towards asking questions based on the previously generated question-answer pairs. The “Chain” in “Chain of RAG” is thus to be interpreted as a chain of question-answer pairs that are iteratively generated. This is unlike the traditional Chain of Thought, wherein a single prompt handles the entire process of creating questions, answers, and follow-up question in one go. Moreover, CoRAG follows a zero-shot approach, i.e.\ the LLM is not provided with any example question-answer pairs to influence the reasoning process. An overview of the process is provided in Algorithm~\ref{alg:corag} as well as Figure \ref{fig:corgatorag}. 

The input to the CoRAG module is the \textit{claim} and \textit{image context} from the multimodal claim generation module (§\ref{multimodal-claim-generation}). The LLM is first prompted to generate a question that is intended to answer an aspect of the claim.  
The generated question is passed to the multimodal evidence retriever (§\ref{multimodal-evidence-retrieval}), which obtains evidence to inform the RAG answer. 
Once the answer is generated, the CoRAG process undergoes a follow-up check (effectively an early termination check). 
The follow-up check prompt (see Appendix \ref{sec:generalprompts}) takes as input the LLM-generated claim as well as all the generated question-answer pair(s), and checks whether enough information has been gathered to answer the claim. If the response from the follow-up check is “True”, it asks a follow-up question. The follow-up question is intended to ask for further information, building on top of the previous question-answer pairs such that the claim can be fully addressed. 

A follow-up check occurs after each question-answer generation step. If the follow-up check prompt finds sufficient evidence in the questions and answers generated up until that point, it terminates and passes the evidence to the veracity prediction and explanation generation module. We also set a constraint of a maximum of six questions, after which the CoRAG process terminates even if it does not have enough evidence for the fact-check. We determined this threshold in preliminary experiments on 80 samples, which indicated that this was the highest number of question-answering steps required for the LLM to obtain enough information to address even the more challenging claims.

\subsubsection{RAG-Augmented Reasoning: Tree of RAG (ToRAG)}
In a similar way to how a traditional Tree of Thought \cite{tree} extends Chain of Thought through branching, Tree of RAG (ToRAG) extends our CoRAG approach by creating question branches at each reasoning step. The best question-answer branch is selected at each step. An overview is provided in Algorithm~\ref{alg:torag} as well as Figure \ref{fig:corgatorag}.

The input to the ToRAG module is the \textit{claim} and \textit{image context} from the multimodal claim generation module (§\ref{multimodal-claim-generation}). Upon receiving this input, the ToRAG approach branches into three, each branch asking a unique question to fact-check the claim. 

\begin{algorithm*}
\scriptsize
\caption{Tree of RAG (ToRAG)}
\begin{algorithmic}[1]
\State \textbf{Input:} Claim $C$, Image Context $I$, Image Captions $IC$
\State $\textbf{BestQAPairs} \gets \text{[]}$ \Comment{Initialize an empty list for best Q-A pairs}
\State $\textbf{Questions} \gets \text{GenerateFirstQuestions}(C, I)$ \Comment{Generates three questions}
\State $counter \gets 0$
\State $followUpNeeded \gets \text{True}$
\While{$counter < no\_of\_steps$ \textbf{and} $followUpNeeded$}
    \State $\textbf{QAPairs} \gets \text{[]}$ \Comment{Initializes an empty list for question-answer pairs}
    \For{$Q$ in $\textbf{Questions}$}
        \If{$\text{QuestionAboutImage}(Q)$}
        \State $A \gets \text{ImageQA}(Q, I, IC)$ \Comment{Using image, question, and captions}
    \Else
        \State $A \gets \text{WebQA}(Q)$ \Comment{Standard evidence retrieval}
    \EndIf
        \State $\textbf{QAPairs}.\text{append}((Q, A))$
    \EndFor
    \State $(\textbf{BestQ}, \textbf{BestA}) \gets \text{QAElimination}(\textbf{QAPairs})$
    \State $\textbf{BestQAPairs}.\text{append}((\textbf{BestQ}, \textbf{BestA}))$ \Comment{Stores the best Q-A pair of this iteration}
    \State $followUpNeeded \gets \text{FollowupCheck}(\textbf{BestQAPairs})$
    \If{$followUpNeeded$}
        \State $\textbf{Questions} \gets \text{GenerateFollowupQuestions}(\textbf{BestQAPairs})$ \Comment{Generates three follow-up questions}
    \Else
        \State \textbf{break}
    \EndIf
    \State $counter \gets counter + 1$
\EndWhile
\State \textbf{return} $\textbf{BestQAPairs}$ \Comment{Returns all collected best Q-A pairs}

\end{algorithmic}
\label{alg:torag}
\end{algorithm*}

Once the three starting questions have been generated, the ToRAG approach uses the evidence retriever (§\ref{multimodal-evidence-retrieval}) to obtain information and generate answers for each question. The three question-answer pairs are then passed into an elimination prompt, from which only one question-answer pair is chosen as candidate evidence. The model is prompted to perform this elimination based on relevance, detail, additional information, and answer confidence (see Appendix \ref{sec:toragprompts}). 

The candidate evidence then serves as the basis for the follow-up question. Three follow-up questions are generated simultaneously based on the candidate evidence. The evidence retriever fetches answers to these questions, and the LLM generates the answers. New candidate evidence is chosen by the elimination prompt and is added to the existing list of candidate evidence. This list, therefore, stores only the best of the three question-answer pairs obtained at each step. 
Upon gathering sufficient information to fact-check the claim as determined by the follow-up check prompt or reaching a maximum of six candidate evidence question-answer pairs, the ToRAG process terminates, and the list of candidate evidence is passed to the veracity prediction and explanation generation module. A few examples of the question-answer pairs generated by our LLM-based and RAG-augmented reasoning approaches can be seen in Appendix \ref{examplesofcoragtorag}.

\subsection{Veracity Prediction and Explanation}

The veracity prediction and explanation module (henceforth referred to as ‘‘veracity prediction'' for brevity) generates a veracity label of \textit{supported} or \textit{refuted} based on the information available in the question-answer pairs. Moreover, it generates a \textit{failed} label when it deems to have insufficient information in the question-answer pair to either support or refute the claim. 

We experiment with three variants of veracity prediction prompts (see Appendix \ref{prompts_veracity}). (i)~The standard veracity prompt (Standard\textsubscript{VP}) takes the claim and evidence pairs as input, and outputs the veracity rating and the explanation without any induced reasoning. (ii)~The zero-shot Chain of Thought veracity prediction prompt (CoT\textsubscript{VP}) uses the “Lets think step by step” phrase to guide the model to follow a chain of thought reasoning approach. (iii)~The Chain of Verification \cite{dhuliawala2023chainofve} veracity prediction prompt (CoVe) first constructs verification questions based on the LLM-generated fact-checked explanation. The answers to these questions are generated using RAG, and are passed -- along with the LLM-generated fact-check -- to a correction check prompt. In case of corrections to the original LLM-generated fact-check, a new fact-check is generated along with a new veracity label if necessary. The CoVe veracity prediction approach is thus able to verify the fact-checked explanation generated by the CoRAG and ToRAG methods with the intended goal of capturing and correcting hallucination.

\newcolumntype{R}[1]{>{\RaggedLeft\arraybackslash}p{#1}}

\begin{table*}[ht!]
  \begin{center}
  \scriptsize
    \resizebox{\textwidth}{!}{%
    \begin{tabular}{|l|R{2cm}|R{2cm}|R{2cm}| R{2cm}|}
      \hline
      \textbf{APPROACHES} & \textbf{SUPPORTED (F1)} & \textbf{REFUTED (F1)} & \textbf{\# FAILED} & \textbf{WEIGHTED F1}\\
      \hline
      SubQ + CoT\textsubscript{VP} &0.66  & 0.77  & 50 $|$ 22 & 0.71\\
      \hline
      CoRAG + Standard\textsubscript{VP} & 0.74 & 0.81 & 31 $|$ 15 & 0.77\\
      CoRAG + CoT\textsubscript{VP} & 0.73  & 0.82  & 38 $|$ 14 & 0.77\\
      CoRAG + CoT\textsubscript{VP} + CoVe & 0.78 & 0.83  & 21 $|$ 8 & 0.81\\
      \hline
      ToRAG + Standard\textsubscript{VP} & 0.82  & \fontseries{b}\selectfont 0.86  & 16 $|$ 5 & 0.84\\
      ToRAG + CoT\textsubscript{VP} & 0.82 & 0.85  & 19 $|$ 9 & 0.83\\
      ToRAG + CoT\textsubscript{VP} + CoVe & \fontseries{b}\selectfont 0.84  & \fontseries{b}\selectfont 0.86  & 9 $|$ 4 & \fontseries{b}\selectfont 0.85\\
      \hline
    \end{tabular}
    }
    \caption{F1 Results of the Correctness of Veracity Predictions evaluation. The \# FAILED column contains the number of \textit{supported $|$ refuted} claims that were predicted as \textit{failed}. 
    }
    \label{tab:defaultmetric1}
  \end{center}
\end{table*}

\section{Evaluation and Results}
We now present two evaluations employed across the set of 300 multimodal claims. In Section \ref{sect_defaultmetric}, we analyze system performance based on the correctness of veracity predictions. 
In Section \ref{expgen}, we zoom into explanation generation by conducting a human annotation study to compare the generated and gold explanations.


\subsection{Correctness of Veracity Predictions} \label{sect_defaultmetric}

\begin{figure*}[ht]
    \centering
    \includegraphics[width=13cm]{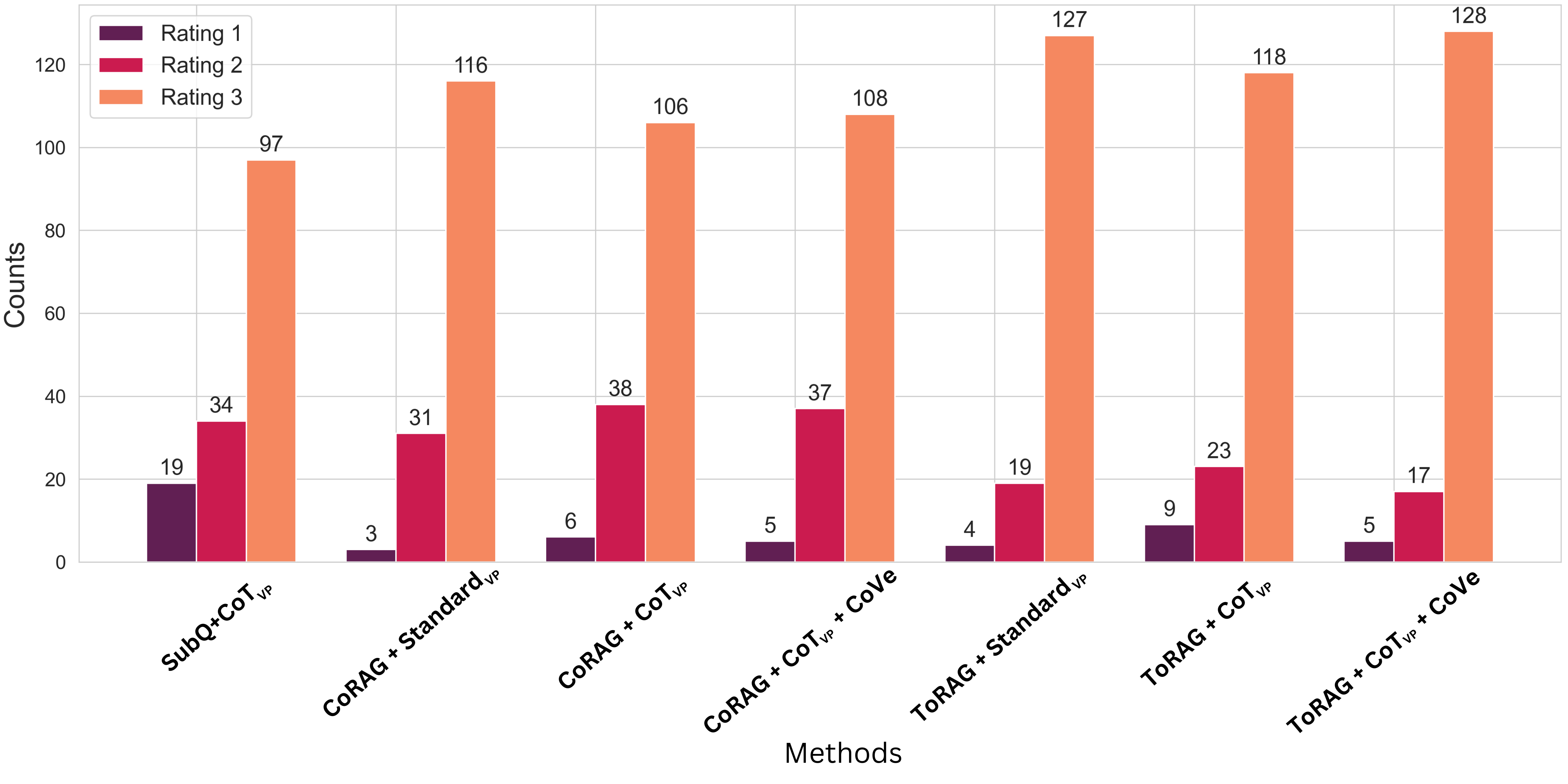}
    \caption{Number of 1/2/3 ratings received for explanations by each approach}
    \label{fig:supportedexplanation}
\end{figure*}

In this evaluation setup, we categorize the predictions into two primary outcomes: correct or incorrect. Specifically, when the language model's prediction matches the actual label (for instance, predicting \textit{supported} when the actual rating is \textit{supported}), the prediction is deemed correct. Conversely, if the model predicts \textit{refuted} or \textit{failed} when the actual rating is \textit{supported}, the prediction is considered as incorrect. Table \ref{tab:defaultmetric1} shows the results of all of our approaches for this evaluation criterion.

The worst-performing approach is the SubQ+CoT\textsubscript{VP} baseline, with a weighted F1 of 0.71.
The best-performing approach is ToRAG+CoT\textsubscript{VP}+CoVe, with a weighted F1 of 0.85.
The middle spot is occupied by the CoRAG implementations; the strongest among those is CoRAG+CoT\textsubscript{VP}+CoVe, with a weighted F1 of 0.81. Regarding class-level performance, the scores are consistently higher for the \textit{refuted} rather than \textit{supported} class.



The SubQ+CoT\textsubscript{VP} baseline lags behind our RAGAR approaches by up to 0.14 weighted F1 points. 
We attribute its poor performance to the inability of the veracity prediction module (CoT\textsubscript{VP}) to gain sequential and contextual information. Since the sub-questions generated by SubQ+CoT\textsubscript{VP} are based solely on the claim, the answers queried during evidence retrieval do not follow from one another. 


Amongst our RAGAR approaches, applying CoT\textsubscript{VP} to the question-answer pairs generated by either CoRAG or ToRAG approaches did not show improvement over Standard\textsubscript{VP}. We attribute this to the very strong internal reasoning capabilities of GPT-4. However, we are able to improve performance by combining the CoVe approach, especially in the case of CoRAG. Incorporating CoVe with the result from CoRAG+CoT\textsubscript{VP} shows a performance improvement of 0.04 F1 points and especially improves the classification of \textit{supported} claims. Incorporating CoVe on top of the ToRAG+CoT\textsubscript{VP} leads to an improvement, but overall minor and also less pronounced than for CoRAG. This indicates that the QA elimination prompt in ToRAG successfully eliminates erroneous or irrelevant question-answer pairs.

\subsection{Evaluating Explanation Generation} \label{expgen}

We evaluate explanation generation by comparing the LLM-generated fact-checked explanation with the corresponding ‘‘Ruling Outline'' from the MOCHEG dataset. We recruit three volunteer annotators, aged 21--24 and with near-native English proficiency. They are asked to rate the explanations generated by each of the approaches on a scale from 1 to 3, where 3 indicates that all information in the gold explanation is present in the generated explanation, while 1 indicates that all information in the gold explanation is missing from the generated explanation. The complete annotation instructions are provided in Appendix~\ref{annotation_section}.

We randomly sample a set of 50 claims, divided into 25 supported and 25 refuted. For all annotated claims, the gold veracity label and the predicted veracity label match. 
We measure inter-annotator agreement using Krippendorff’s $\alpha$ \citep{Hayes2007AnsweringTC}. The scores are in the range of 0.53 to 0.75 depending on the evaluated approach, with the mean at 0.60. We consider this to be sufficient agreement given the nature of the task. 



As can be seen in Figure \ref{fig:supportedexplanation}, the annotators provide a rating of 3 for an overwhelming majority of explanations generated across methods. This shows that the generated explanations indeed cover all the points noted in the PolitiFact fact-check. 
Additionally, the explanations generated by SubQ+CoT\textsubscript{VP} led to significantly more ratings of 1 than any other method, which indicates that it omitted or did not accurately elaborate on certain points. 

Regarding class-level trends, explanations in the \textit{supported} class are rated as 2 more often than those in the \textit{refuted} class (see Appendix \ref{extra_explanation_graph}). This indicates that certain information was missing from the generated explanation; more generally, this trend reflects the lower F1 scores on this class (§\ref{sect_defaultmetric}), suggesting its higher difficulty. From a qualitative perspective, the annotators anecdotally reported that the generated explanations included some points from the PolitiFact ruling outline, but also provided additional information.
Overall, however, the majority of the ratings being annotated as 3 across the different approaches lends credence to the quality of the explanation and to the efficacy of the underlying system in retrieving relevant evidence to fact-check the claim.

\section{Conclusion}
This paper introduces and tests two new methods for political fact-checking using large language models (LLMs): Chain of RAG (CoRAG) and Tree of RAG (ToRAG). These methods tackle misinformation in political discussions, focusing on multimodal claims, and show notable improvements over traditional fact-checking approaches that use sub-question generation with LLMs. CoRAG uses a step-by-step questioning strategy for thorough claim examination, while ToRAG extends upon this by following a branching strategy with evidence elimination thereby enhancing veracity prediction. We evaluate these methods in two ways. In terms of correctness of generated veracity label, we see an increase of 0.06-0.14 F1 points when using the RAGAR framework with Standard, CoT\textsubscript{VP}, and CoVe veracity prediction prompts compared to the baseline SubQ+CoT\textsubscript{VP}. For explanation generation, the quality of RAGAR-generated explanations was consistently rated higher than the baseline method. Our study shows that RAG-augmented reasoning (RAGAR) techniques are effective in multimodal political fact-checking, improving both the accuracy of veracity predictions and the quality of detailed fact-check explanations.

\section{Limitations}

We experimented with three tools for extracting relevant web results for natural language questions; DuckDuckGo Search, You.com\footnote{\url{https://you.com/}} and Tavily AI\footnote{\url{https://tavily.com/}}. Across the three tools, we notice that the search results may occasionally vary when prompted with the same questions multiple times. This variance in results, even though the question remains the same or similar, is problematic since it affects the final result and makes it hard to compare approaches. Additionally, due to budget constraints, we are unable to provide variance estimates requiring multiple runs of our RAGAR approaches. While we acknowledge the use of a closed-source LLM as a potential shortcoming due to comparatively more limited control over model behavior, we opted for the best-performing model available to us given the complexity of the addressed task.
Finally, as also noted in the paper, our main aim was to assess the viability of novel reasoning techniques rather than retrieval quality, which led us to exclude \textit{NEI} instances from our experimental setup. Further work extended to these cases is needed to more comprehensively understand the performance of our proposed approach.

\section{Ethics Statement}
We conducted an experimental study aimed at examining multimodal fact-checking by prompting LLMs, 
and note that some of the core steps of this approach may also be replicated by the general public.
Our RAGAR approach obtained clear improvements over the examined baseline in the evaluation setup we defined.
However, the experiments presented here are not sufficient to make general claims about the performance of our approach in other settings.
Given the sensitive nature of political news in particular, we caution
against using the RAGAR approach for general political fact-checking or implementing it on a large scale at this stage.

\bibliography{ACL/custom}

\begin{thebibliography}{23}
\expandafter\ifx\csname natexlab\endcsname\relax\def\natexlab#1{#1}\fi

\bibitem[{Alam et~al.(2022)Alam, Cresci, Chakraborty, Silvestri, Dimitrov, Martino, Shaar, Firooz, and Nakov}]{alam-etal-2022-survey}
Firoj Alam, Stefano Cresci, Tanmoy Chakraborty, Fabrizio Silvestri, Dimiter Dimitrov, Giovanni Da~San Martino, Shaden Shaar, Hamed Firooz, and Preslav Nakov. 2022.
\newblock \href {https://aclanthology.org/2022.coling-1.576} {A survey on multimodal disinformation detection}.
\newblock In \emph{Proceedings of the 29th International Conference on Computational Linguistics}, pages 6625--6643, Gyeongju, Republic of Korea. International Committee on Computational Linguistics.

\bibitem[{Asai et~al.(2024)Asai, Wu, Wang, Sil, and Hajishirzi}]{asai2024selfrag}
Akari Asai, Zeqiu Wu, Yizhong Wang, Avirup Sil, and Hannaneh Hajishirzi. 2024.
\newblock \href {https://openreview.net/forum?id=hSyW5go0v8} {Self-{RAG}: Learning to retrieve, generate, and critique through self-reflection}.
\newblock In \emph{The Twelfth International Conference on Learning Representations}.

\bibitem[{Aïmeur et~al.(2023)Aïmeur, Amri, and Brassard}]{aimeur_fake_2023}
Esma Aïmeur, Sabrine Amri, and Gilles Brassard. 2023.
\newblock \href {https://doi.org/10.1007/s13278-023-01028-5} {Fake news, disinformation and misinformation in social media: a review}.
\newblock \emph{Social Network Analysis and Mining}, 13(1):30.

\bibitem[{Chern et~al.(2023)Chern, Chern, Chen, Yuan, Feng, Zhou, He, Neubig, and Liu}]{chern2023factool}
I-Chun Chern, Steffi Chern, Shiqi Chen, Weizhe Yuan, Kehua Feng, Chunting Zhou, Junxian He, Graham Neubig, and Pengfei Liu. 2023.
\newblock \href {http://arxiv.org/abs/2307.13528} {{FacTool}: Factuality detection in generative {AI} -- a tool augmented framework for multi-task and multi-domain scenarios}.
\newblock \emph{arXiv preprint arXiv:2307.13528}.

\bibitem[{Das et~al.(2023)Das, Liu, Kovatchev, and Lease}]{DAS2023103219}
Anubrata Das, Houjiang Liu, Venelin Kovatchev, and Matthew Lease. 2023.
\newblock \href {https://doi.org/https://doi.org/10.1016/j.ipm.2022.103219} {The state of human-centered {NLP} technology for fact-checking}.
\newblock \emph{Information Processing \& Management}, 60(2):103219.

\bibitem[{Dhuliawala et~al.(2023)Dhuliawala, Komeili, Xu, Raileanu, Li, Celikyilmaz, and Weston}]{dhuliawala2023chainofve}
Shehzaad Dhuliawala, Mojtaba Komeili, Jing Xu, Roberta Raileanu, Xian Li, Asli Celikyilmaz, and Jason Weston. 2023.
\newblock Chain-of-verification reduces hallucination in large language models.
\newblock \emph{arXiv preprint arXiv:2309.11495}.

\bibitem[{Guo et~al.(2023)Guo, Dong, Ji, Bai, Guo, and Zuo}]{guo2023texts}
Zixian Guo, Bowen Dong, Zhilong Ji, Jinfeng Bai, Yiwen Guo, and Wangmeng Zuo. 2023.
\newblock Texts as images in prompt tuning for multi-label image recognition.
\newblock In \emph{Proceedings of the IEEE/CVF Conference on Computer Vision and Pattern Recognition}, pages 2808--2817.

\bibitem[{Hayes and Krippendorff(2007)}]{Hayes2007AnsweringTC}
Andrew~F. Hayes and Klaus Krippendorff. 2007.
\newblock Answering the call for a standard reliability measure for coding data.
\newblock \emph{Communication Methods and Measures}, 1:77 -- 89.

\bibitem[{Lewis et~al.(2020)Lewis, Liu, Goyal, Ghazvininejad, Mohamed, Levy, Stoyanov, and Zettlemoyer}]{lewis2019bart}
Mike Lewis, Yinhan Liu, Naman Goyal, Marjan Ghazvininejad, Abdelrahman Mohamed, Omer Levy, Veselin Stoyanov, and Luke Zettlemoyer. 2020.
\newblock \href {https://aclanthology.org/2020.acl-main.703} {{BART}: Denoising sequence-to-sequence pre-training for natural language generation, translation, and comprehension}.
\newblock In \emph{Proceedings of the 58th Annual Meeting of the Association for Computational Linguistics}, pages 7871--7880, Online. Association for Computational Linguistics.

\bibitem[{OpenAI(2023)}]{openai2023gpt4v}
OpenAI. 2023.
\newblock \href {https://cdn.openai.com/contributions/gpt-4v.pdf} {Gpt-4v: A multimodal transformer for vision and language}.

\bibitem[{Pan et~al.(2023)Pan, Wu, Lu, Luu, Wang, Kan, and Nakov}]{pan-etal-2023-fact}
Liangming Pan, Xiaobao Wu, Xinyuan Lu, Anh~Tuan Luu, William~Yang Wang, Min-Yen Kan, and Preslav Nakov. 2023.
\newblock \href {https://aclanthology.org/2023.acl-long.386} {Fact-checking complex claims with program-guided reasoning}.
\newblock In \emph{Proceedings of the 61st Annual Meeting of the Association for Computational Linguistics (Volume 1: Long Papers)}, pages 6981--7004, Toronto, Canada. Association for Computational Linguistics.

\bibitem[{Pan et~al.(2024)Pan, Luo, Li, and Liu}]{chainofaction}
Zhenyu Pan, Haozheng Luo, Manling Li, and Han Liu. 2024.
\newblock Chain-of-action: Faithful and multimodal question answering through large language models.
\newblock \emph{arXiv preprint arXiv:2403.17359}.

\bibitem[{Peng et~al.(2023)Peng, Galley, He, Cheng, Xie, Hu, Huang, Liden, Yu, Chen, and Gao}]{peng2023check}
Baolin Peng, Michel Galley, Pengcheng He, Hao Cheng, Yujia Xie, Yu~Hu, Qiuyuan Huang, Lars Liden, Zhou Yu, Weizhu Chen, and Jianfeng Gao. 2023.
\newblock \href {http://arxiv.org/abs/2302.12813} {Check your facts and try again: Improving large language models with external knowledge and automated feedback}.
\newblock \emph{arXiv preprint arXiv:2302.12813}.

\bibitem[{Radford et~al.(2021)Radford, Kim, Hallacy, Ramesh, Goh, Agarwal, Sastry, Askell, Mishkin, Clark, Krueger, and Sutskever}]{radford2021learning}
Alec Radford, Jong~Wook Kim, Chris Hallacy, Aditya Ramesh, Gabriel Goh, Sandhini Agarwal, Girish Sastry, Amanda Askell, Pamela Mishkin, Jack Clark, Gretchen Krueger, and Ilya Sutskever. 2021.
\newblock \href {https://proceedings.mlr.press/v139/radford21a.html} {Learning transferable visual models from natural language supervision}.
\newblock In \emph{Proceedings of the 38th International Conference on Machine Learning}, volume 139 of \emph{Proceedings of Machine Learning Research}, pages 8748--8763. PMLR.

\bibitem[{Vo and Lee(2019)}]{vo2019learning}
Nguyen Vo and Kyumin Lee. 2019.
\newblock \href {https://doi.org/10.1145/3331184.3331248} {Learning from fact-checkers: Analysis and generation of fact-checking language}.
\newblock In \emph{Proceedings of the 42nd International ACM SIGIR Conference on Research and Development in Information Retrieval}, SIGIR'19, page 335–344, New York, NY, USA. Association for Computing Machinery.

\bibitem[{Vosoughi et~al.(2018)Vosoughi, Roy, and Aral}]{vosoughi2018}
Soroush Vosoughi, Deb Roy, and Sinan Aral. 2018.
\newblock \href {http://arxiv.org/abs/https://www.science.org/doi/pdf/10.1126/science.aap9559} {The spread of true and false news online}.
\newblock \emph{Science}, 359(6380):1146--1151.

\bibitem[{Wei et~al.(2022)Wei, Wang, Schuurmans, Bosma, Ichter, Xia, Chi, Le, and Zhou}]{chainofthought}
Jason Wei, Xuezhi Wang, Dale Schuurmans, Maarten Bosma, Brian Ichter, Fei Xia, Ed~Chi, Quoc~V Le, and Denny Zhou. 2022.
\newblock Chain-of-thought prompting elicits reasoning in large language models.
\newblock In \emph{Advances in Neural Information Processing Systems}, volume~35, pages 24824--24837. Curran Associates, Inc.

\bibitem[{Xu et~al.(2023)Xu, Pang, Shen, Cheng, and Chua}]{xu2023search}
Shicheng Xu, Liang Pang, Huawei Shen, Xueqi Cheng, and Tat-seng Chua. 2023.
\newblock Search-in-the-chain: Towards the accurate, credible and traceable content generation for complex knowledge-intensive tasks.
\newblock \emph{arXiv preprint arXiv:2304.14732}.

\bibitem[{Yao et~al.(2023{\natexlab{a}})Yao, Shah, Sun, Cho, and Huang}]{Yao_2023}
Barry~Menglong Yao, Aditya Shah, Lichao Sun, Jin-Hee Cho, and Lifu Huang. 2023{\natexlab{a}}.
\newblock \href {https://doi.org/10.1145/3539618.3591879} {End-to-end multimodal fact-checking and explanation generation: A challenging dataset and models}.
\newblock In \emph{Proceedings of the 46th International {ACM} {SIGIR} Conference on Research and Development in Information Retrieval}. {ACM}.

\bibitem[{Yao et~al.(2023{\natexlab{b}})Yao, Yu, Zhao, Shafran, Griffiths, Cao, and Narasimhan}]{tree}
Shunyu Yao, Dian Yu, Jeffrey Zhao, Izhak Shafran, Thomas~L. Griffiths, Yuan Cao, and Karthik Narasimhan. 2023{\natexlab{b}}.
\newblock \href {http://arxiv.org/abs/2305.10601} {Tree of thoughts: Deliberate problem solving with large language models}.
\newblock \emph{arXiv preprint arXiv:2305.10601}.

\bibitem[{Zannettou et~al.(2018)Zannettou, Caulfield, Blackburn, De~Cristofaro, Sirivianos, Stringhini, and Suarez-Tangil}]{zannettou2018origins}
Savvas Zannettou, Tristan Caulfield, Jeremy Blackburn, Emiliano De~Cristofaro, Michael Sirivianos, Gianluca Stringhini, and Guillermo Suarez-Tangil. 2018.
\newblock \href {https://doi.org/10.1145/3278532.3278550} {On the origins of memes by means of fringe web communities}.
\newblock In \emph{Proceedings of the Internet Measurement Conference 2018}, IMC '18, page 188–202, New York, NY, USA. Association for Computing Machinery.

\bibitem[{Zeng and Gao(2024)}]{Zeng2024JustiLMFJ}
Fengzhu Zeng and Wei Gao. 2024.
\newblock Justilm: Few-shot justification generation for explainable fact-checking of real-world claims.
\newblock \emph{arXiv preprint arXiv:2401.08026}.

\bibitem[{Zhang and Gao(2023)}]{zhang2023llmbased}
Xuan Zhang and Wei Gao. 2023.
\newblock Towards llm-based fact verification on news claims with a hierarchical step-by-step prompting method.
\newblock \emph{arXiv preprint arXiv:2310.00305}.

\end{thebibliography}
\bibliographystyle{acl_natbib}

\clearpage

\appendix
\onecolumn

\section{Appendix}
\label{sec:appendix}

\subsection{Instructions to Annotators}
\label{annotation_section}
\begin{figure*}[ht]
    \centering
    \includegraphics[width=11cm]{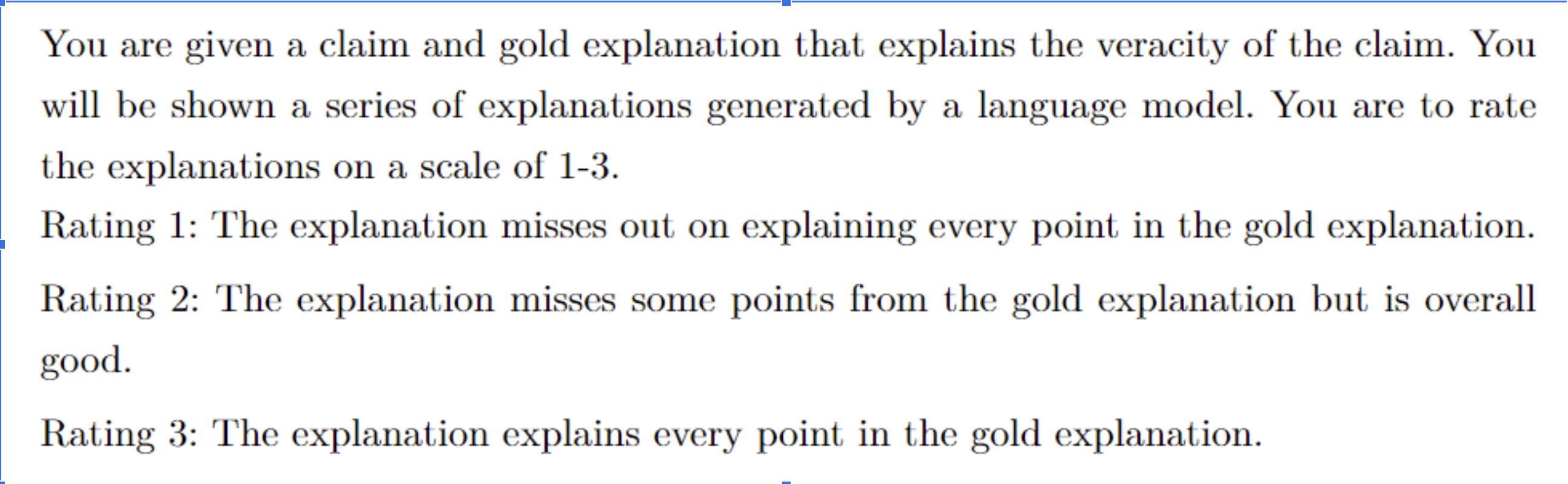}
    \caption{Annotation Instructions}
    \label{fig:annotation}
\end{figure*}

The instructions to annotators for the evaluation of the Explanation Generation Task is provided in Figure \ref{fig:annotation}.

\subsection{Explanation Generation by Veracity Label}
\label{extra_explanation_graph}
\begin{figure*}[ht]
    \centering
    \includegraphics[width=11.5cm]{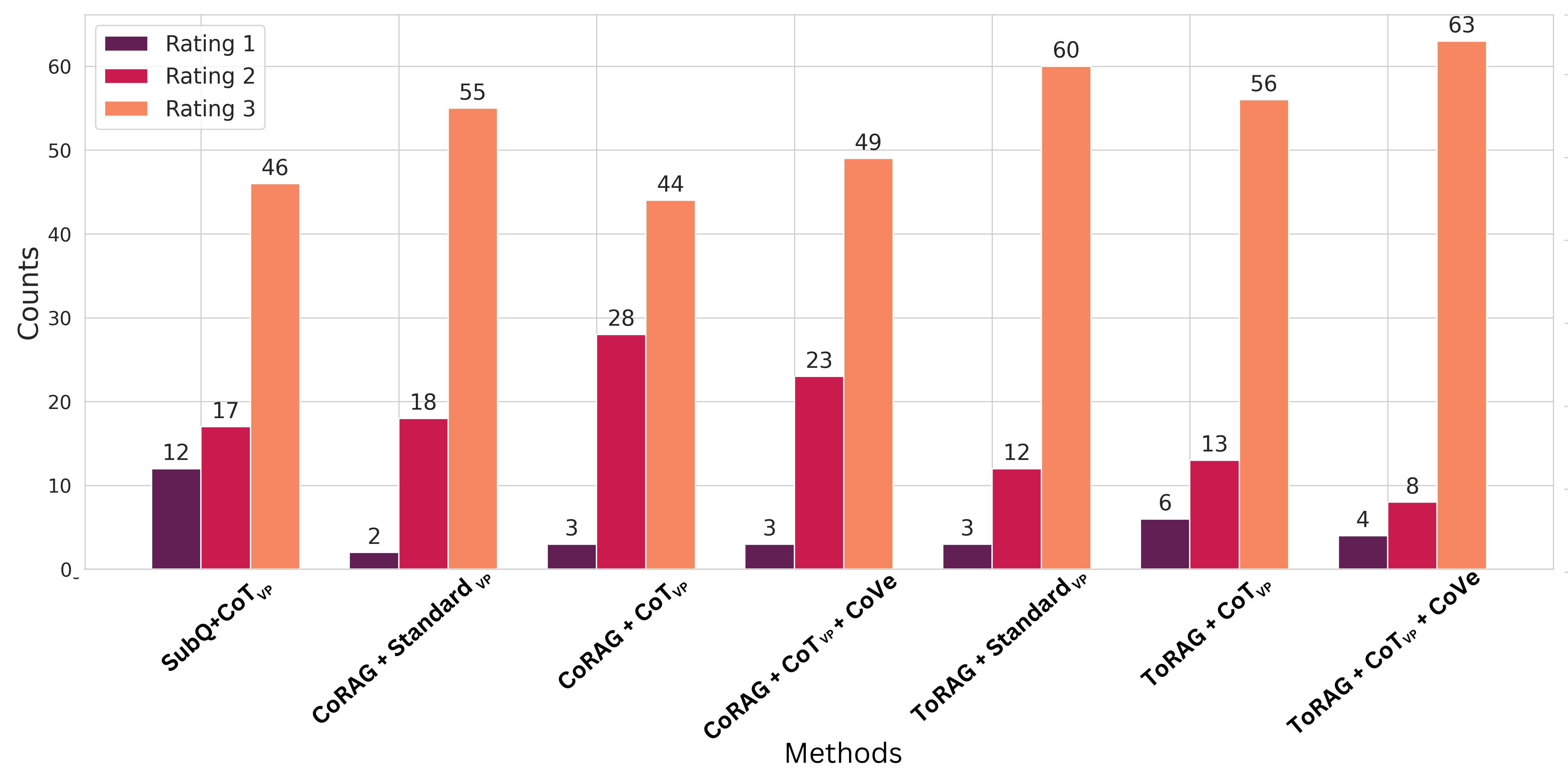}
    \caption{Annotator ratings for explanations of supported claims}
    \label{fig:supportedgraph}
\end{figure*}

\begin{figure*}[ht]
    \centering
    \includegraphics[width=11.5cm]{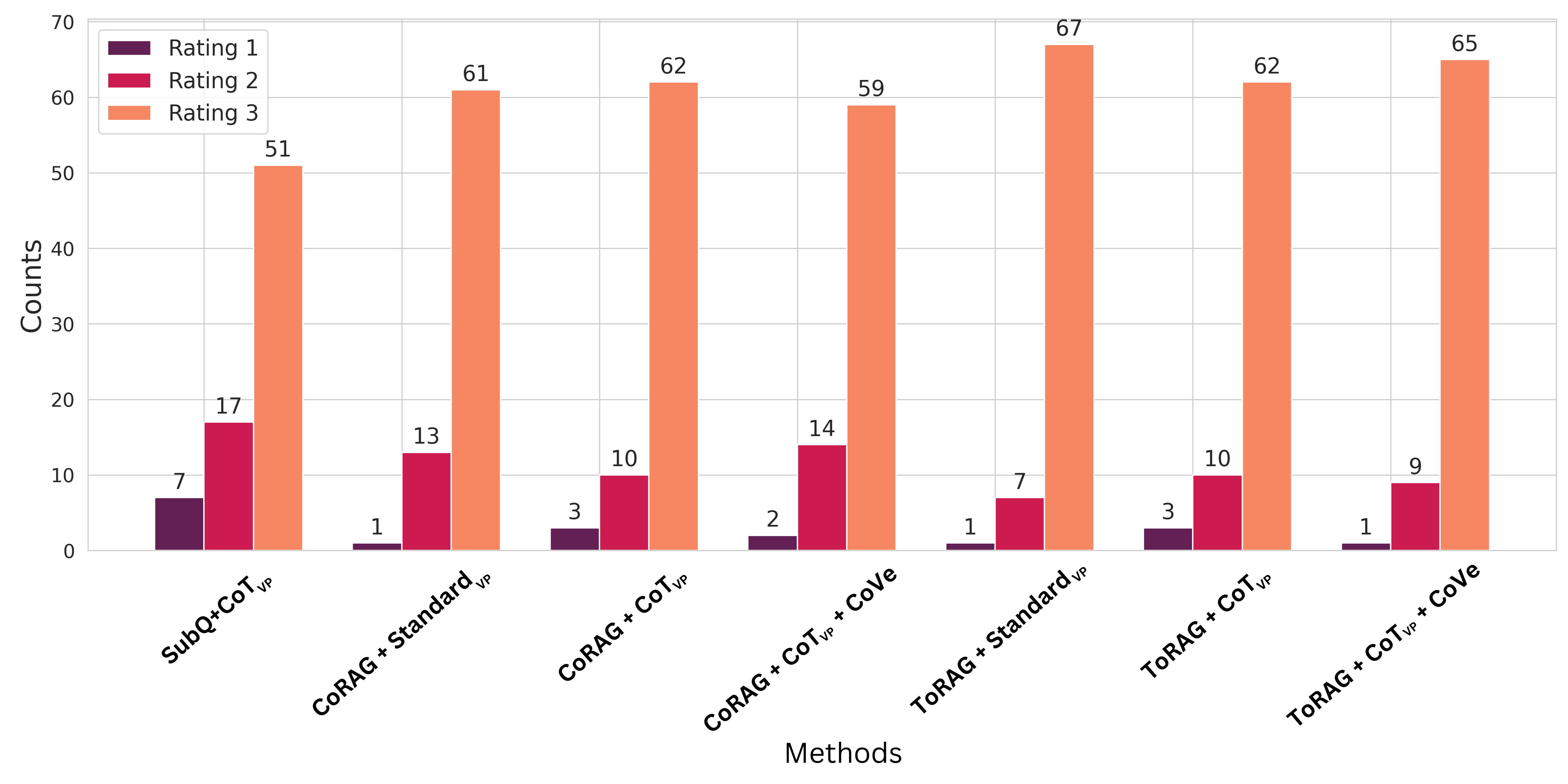}
    \caption{Annotator ratings for explanations of refuted claims}
    \label{fig:refutedgraph}
\end{figure*}

In addition to the overall ratings for the Human Annotation for Explanation Generation, we also provide the ratings for specific classes. Figure \ref{fig:supportedgraph} shows the human annotation ratings for the explanations of supported claims. Figure \ref{fig:refutedgraph} shows the human annotation ratings for the explanations of refuted claims.


\subsection{Discussing Multimodal RAG}
\label{multimodalragdisc}
We utilize reverse image search to extract captions of matching images from the web. We showcase the Image QA pairs for the examples in Table \ref{mmdiscussion}. The first example regarding Mike Pompeo showcases how GPT-4V is unable to identify the Afghan dignitary and the image context is unable to provide a name that could help fact-check the claim. However, using the image captions retrieved from the internet and prompting the evidence retrieval along with the image caption, GPT-4V is able to identify the Afghan dignitary as Mullah Abdul Ghani Baradar. The fact-check then continues to verify if Mullah Abdul Ghani Baradar was indeed ever the Afghan President or not. Similarly, in the third example with Joe Biden kneeling, the image captions extracted by reverse image search are able to add the additional information that Joe Biden was kneeling down to pose with dancers in Haiti. This information is crucial for the particular fact-check since it contextualizes the reason why Joe Biden was kneeling as well as detailing the event where the described act occurred.

\begin{longtable}{|>{\raggedright\arraybackslash}p{0.2\textwidth}p{0.20\textwidth}p{0.25\textwidth}p{0.22\textwidth}|}
\caption{Example table with claims, images, and QA.} 
\label{mmdiscussion}
\\
\hline
 \textbf{Claim} & \textbf{Image} & \textbf{Generated Image Context} & \textbf{Image QA} \\
 \hline
\endfirsthead

\multicolumn{4}{c}%
{{\bfseries \tablename\ \thetable{} -- continued from previous page}} \\
\hline
 \textbf{Claim} & \textbf{Image} & \textbf{Claim Generation} & \textbf{Image QA} \\
 \hline
\endhead

\hline
\endfoot

\hline
\endlastfoot

\footnotesize The man next to Mike Pompeo in a November 2020 photo is the guy the Trump administration helped get out of jail in 2018 and who is now the 'president' of Afghanistan. & \raisebox{-\totalheight}{\includegraphics[width=0.20\textwidth,keepaspectratio]{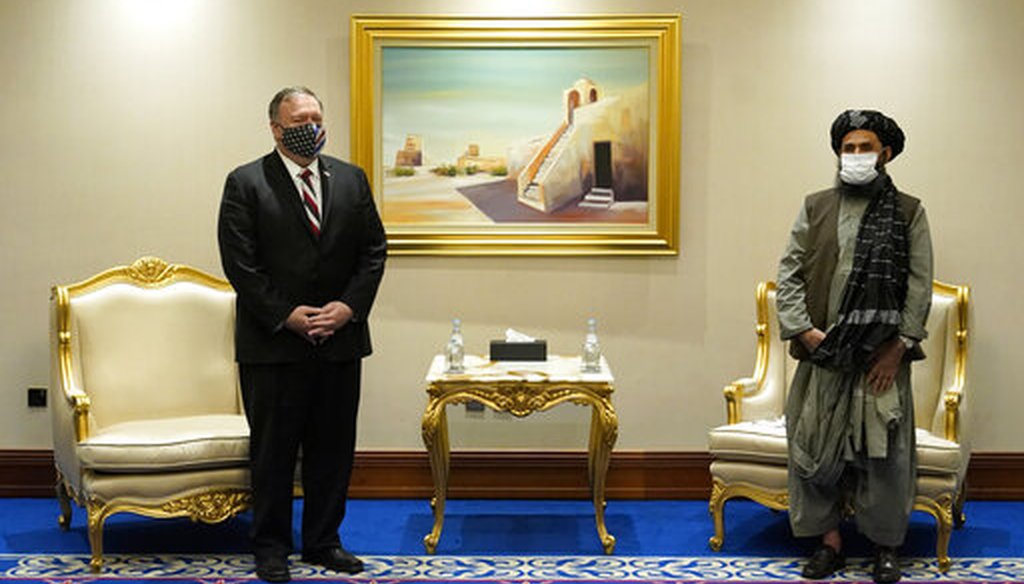}} & \scriptsize The image shows Mike Pompeo standing next to a man wearing traditional Afghan attire and a face mask. The setting appears to be a formal meeting room with a painting on the wall, two gold-colored chairs, and a table with water bottles. Both individuals are standing apart, likely adhering to social distancing guidelines prevalent during the COVID-19 pandemic. The background and attire suggest the meeting is official in nature and likely related to diplomatic or governmental affairs. Claim stated in 2021. & \scriptsize Q. Who is the man standing next to Mike Pompeo in the November 2020 photo? 

\scriptsize A. Mullah Abdul Ghani Baradar \\
\hline

\footnotesize Says Joe Biden took a knee during the national anthem.
 & \raisebox{-\totalheight}{\includegraphics[width=0.20\textwidth,keepaspectratio]{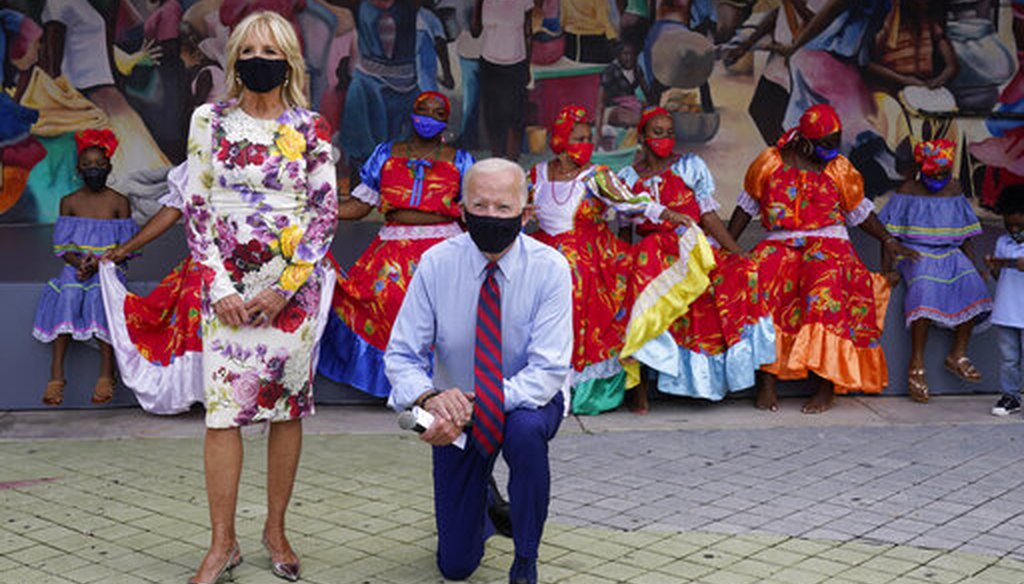}} & \scriptsize The image shows Joe Biden taking a knee, but it does not provide evidence that this act occurred during the national anthem. Without additional context, the claim cannot be confirmed solely based on this image. The time frame of the claim is 'Claim stated in 2020'.
 & \scriptsize Is there a specific date and location associated with the image of Joe Biden taking a knee? 

 \scriptsize A. FILE - In this Monday, Oct. 5, 2020 file photo, Democratic presidential candidate former Vice President Joe Biden and his wife Jill Biden pose for a photo with dancers as they visit Little Haiti Cultural Complex in Miami. \\
\hline
 
\hline

\footnotesize The Trump administration worked to free 5,000 Taliban prisoners.
 & \raisebox{-\totalheight}{\includegraphics[width=0.20\textwidth,keepaspectratio]{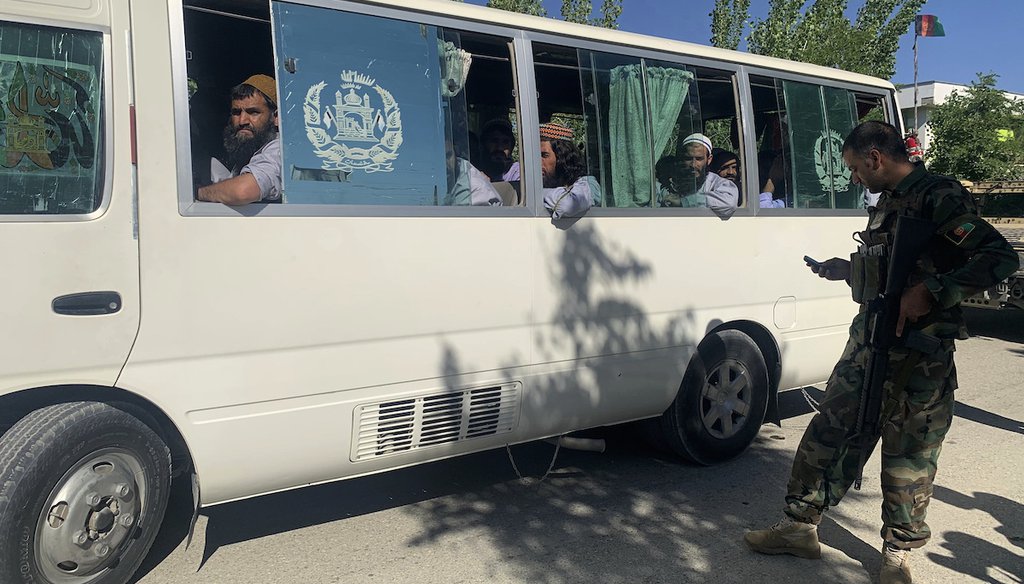}} & \scriptsize The image shows individuals, presumed to be Taliban prisoners, inside a bus with a guard standing nearby, which potentially correlates to the release of Taliban prisoners. The context suggests this may represent a prisoner release process. & \scriptsize Q. Were the individuals shown in the provided image actually Taliban prisoners being released as part of the agreement? 

\scriptsize A.'Taliban prisoners are released from Pul-e-Charkhi jail in Kabul, Afghanistan, Thursday, Aug. 13, 2020 \\

\hline
\footnotesize These were not chemical irritants' used to clear a crowd. Pepper balls are 'not a chemical irritant.
 & \raisebox{-\totalheight}{\includegraphics[width=0.20\textwidth,keepaspectratio]{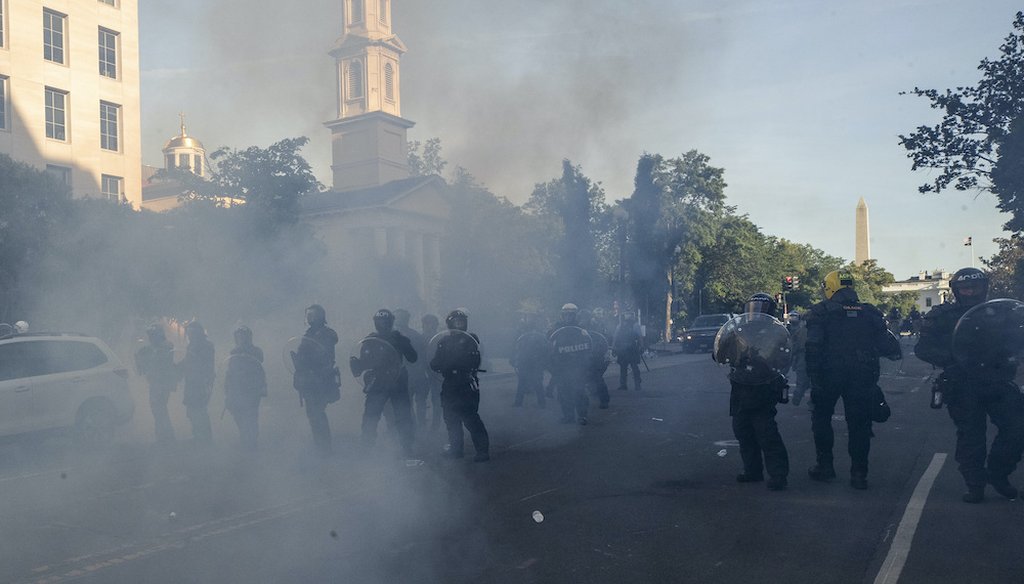}} & \scriptsize The image shows law enforcement in protective gear amidst a haze that is consistent with the use of some form of crowd control substance, such as a chemical irritant. Visible smoke and the dispersing crowd strongly suggest the use of a substance to clear the area, counter to the claim that no chemical irritants were used. The presence of pepper balls would depend on identifying specific items or equipment in the scene that are known to dispense pepper balls.

 & \scriptsize  Can we identify the specific equipment or methods used by law enforcement in the image to determine if pepper balls or another substance was deployed? 

\scriptsize A. ... The caption snippets suggest that teargas was used to clear Lafayette Park for a photo opportunity... \\
\hline

\end{longtable}

\clearpage
\subsection{Example Questions}
\label{examplesofcoragtorag}

In Table \ref{exmaplecottoragcorag} and Table \ref{examplecoragtorag} we show the questions asked by each approach.
 
\begin{longtable}{|p{0.15\textwidth}|p{0.80\textwidth}|}
\caption{Example of Question-Answers asked by SubQ+CoT\textsubscript{VP}, CoRAG and ToRAG}
\label{exmaplecottoragcorag} 

\\
\hline
\multicolumn{2}{|c|}{\textbf{Claim:} PPP on average provided a grant of around \$11,000 per employee} \\
\hline
\endfirsthead

\hline
\hline
\multicolumn{2}{|c|}{\textbf{Claim:} PPP on average provided a grant of around \$11,000 per employee} \\
\endhead

\hline
\endfoot

\hline
\endlastfoot

\textbf{SubQ+ CoT\textsubscript{VP}} &\footnotesize  \textbf{Q. What was the total amount of funds disbursed through the Paycheck Protection Program (PPP) in 2020?}

A.More than \$511 billion

\textbf{Q. What criteria were used to determine the amount of grant each business received per employee under the PPP?"}

A. ... determined based on the business's average monthly payroll costs...  approximately 2.5 times the applicant's average monthly payroll... only up to \$100,000 of an employee's annual compensation was eligible for calculation.

\textbf{Q. Were there any variations in the grant amount per employee based on the industry, location, or size of the business?} 

A. ... mentions different aspects of grants, such as administration costs, the impact of public money on grant receivers, and the employee retention credit, which suggests that grant amounts...
\newline
\textbf{Q. How does the PPP define an "employee" for the purposes of calculating grants?}

A. ...means individuals employed on a full-time, part-time, or other basis. \\
\hline
\textbf{CoRAG} &\footnotesize \textbf{Q. What was the average grant amount per employee provided by the Paycheck Protection Program in 2020?}

A. ...for self-employed individuals and employees, the PPP loans could cover up to \$100,000 of annualized pay per employee for eight weeks, with a maximum of \$15,385 per individual."

\textbf{Q. How does the maximum allowable amount of \$15,385 per individual under PPP loans relate to the average grant of around \$11,000 per employee?}

A. The maximum allowable amount of \$15,385 per individual under PPP loans is derived from the limit set for self-employed individuals and employees, which is based on up to \$100,000 of annualized pay for eight weeks...On the other hand, the average grant of around \$11,000 per employee, as observed by Square Capital, reflects the actual average loan amount disbursed to small businesses through the PPP. \\
\hline

\textbf{ToRAG} &\footnotesize \textbf{Q. What was the typical grant amount per employee provided by the PPP as of mid-2020?}

A. ... the average size of a PPP loan approved during the first wave was \$206,000, with 74\% of the loans approved being for amounts of \$150,000 or less."
\newline
\textbf{Q. What is the average monthly payroll cost per employee, as used in the PPP loan calculations, and how does this figure relate to the reported average grant of around \$11,000 per employee?}

A. ...For a 24-week period, the maximum amount per individual is \$46,154, or for an eight-week period, it is \$15,385.
\newline
\textbf{Q. Can you provide the total number of employees covered by PPP loans up to mid-2020 and the total amount of PPP funds disbursed by that time, to calculate the actual average grant per employee?}

A. ... average grant per employee can be calculated as \$511 billion divided by 51 million employees, which equals approximately \$10,019.61 per employee.\\
\hline

\end{longtable}

\begin{longtable}{|p{0.1\textwidth}|p{0.85\textwidth}|}
\caption{Example of Question-Answers asked by CoRAG and ToRAG} 
\label{examplecoragtorag}
\\

\hline
\multicolumn{2}{|c|}{\textbf{Claim:} These 11 mass shootings involved an AR-15-style weapon.} \\
\hline
\endfirsthead

\hline
\multicolumn{2}{|c|}{\textbf{Claim:} (Continued)} \\
\endhead

\hline
\endfoot

\hline
\endlastfoot

\textbf{CoRAG} &\footnotesize \textbf{Q.Were AR-15-style weapons used in 11 mass shootings in the United States before 2021?}

A. …confirms that AR-15-style rifles were used in the 2012 Aurora, Colorado...

\textbf{Q. Can you list the specific mass shootings, aside from the 2012 Aurora, Colorado incident, where AR-15-style rifles were reportedly used before 2021?}

A. ... search results provided do not list specific mass shootings... stating that AR-15 rifles were used in 26 percent of the last 80 mass shootings...before 2021.

\textbf{Q. Given the information that AR-15-style rifles were used in 26 percent of the last 80 mass shootings before 2021, identify other incidents where these rifles were reportedly used.}

A. ... occurred when a gunman, armed with an AR-15-style rifle and body armor, killed nine people and wounded 27 others.

\textbf{Q. Can you identify the specific characteristics or features that define an AR-15-style rifle as used in these incidents?}

A. 1. Chambering …2. Controls 3. Fast and easy reloading…
\\ \hline

\textbf{ToRAG} &\footnotesize \textbf{Q. What are the specific dates and locations of the 11 mass shootings mentioned as involving an AR-15-style weapon?}

A. 1. Boulder, Colorado: March 22, 2021...  2. San Jose, California: May 26, 2021..."

\textbf{Q. Were there any official reports or statements from law enforcement agencies confirming the use of AR-15-style weapons in the Boulder, Colorado, and San Jose, California shootings?}

A. Yes, there were official reports confirming the use of an AR-15-style weapon in the Boulder, Colorado shooting…

\textbf{Q. Can you provide details on the legal acquisition and ownership status of AR-15-style weapons by the shooters in the remaining nine mass shootings mentioned?}

A.
1. Sutherland Springs church shooting: ... goods retailer violated the law ...
2. Boulder supermarket shooting: ... legally purchased the AR-15-style rifle ...
4. Pittsburgh synagogue shooting: ... like the AR-15 rifle used in the attack. 5. Las Vegas shooting: ... claim for selling AR-15s... 
6. Orlando nightclub shooting: ... 
7. San Bernardino shooting: ...
8. Sandy Hook Elementary School shooting: ..."
 \\
\hline

\end{longtable}

\subsection{General Prompts in the RAGAR Approaches}
\label{sec:generalprompts}

\begin{figure}[ht]
    \centering
    \includegraphics[width=14cm]{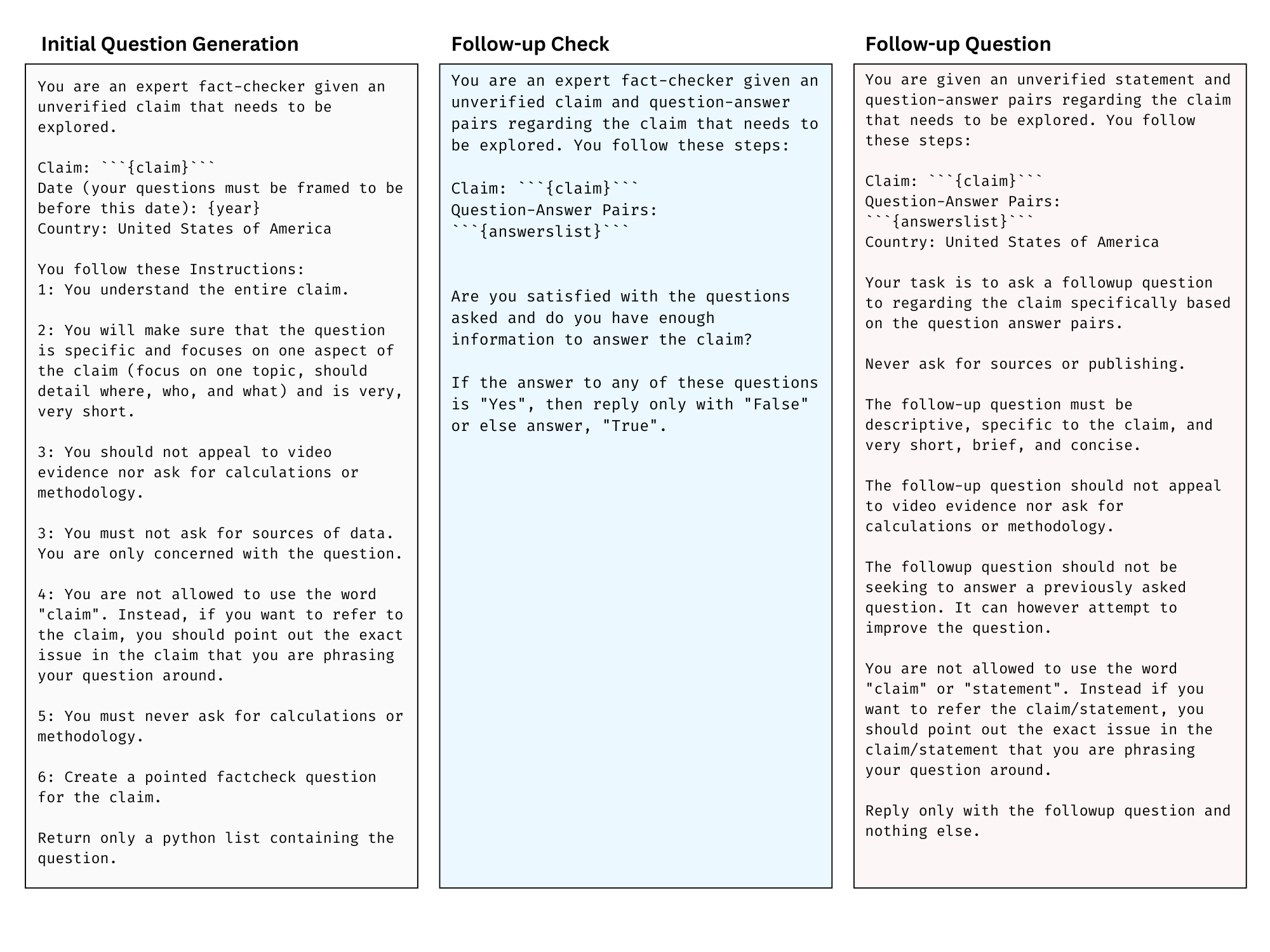}
    \caption{Prompt for initial question-generation, Follow-up Check and Follow-up Question common to all RAGAR approaches}
    \label{fig:questiongeneration}
\end{figure}

\clearpage

\subsection{Prompts Specific to Tree of RAG}
\label{sec:toragprompts}

\begin{figure}[ht]
    \centering
    \includegraphics[width=14cm]{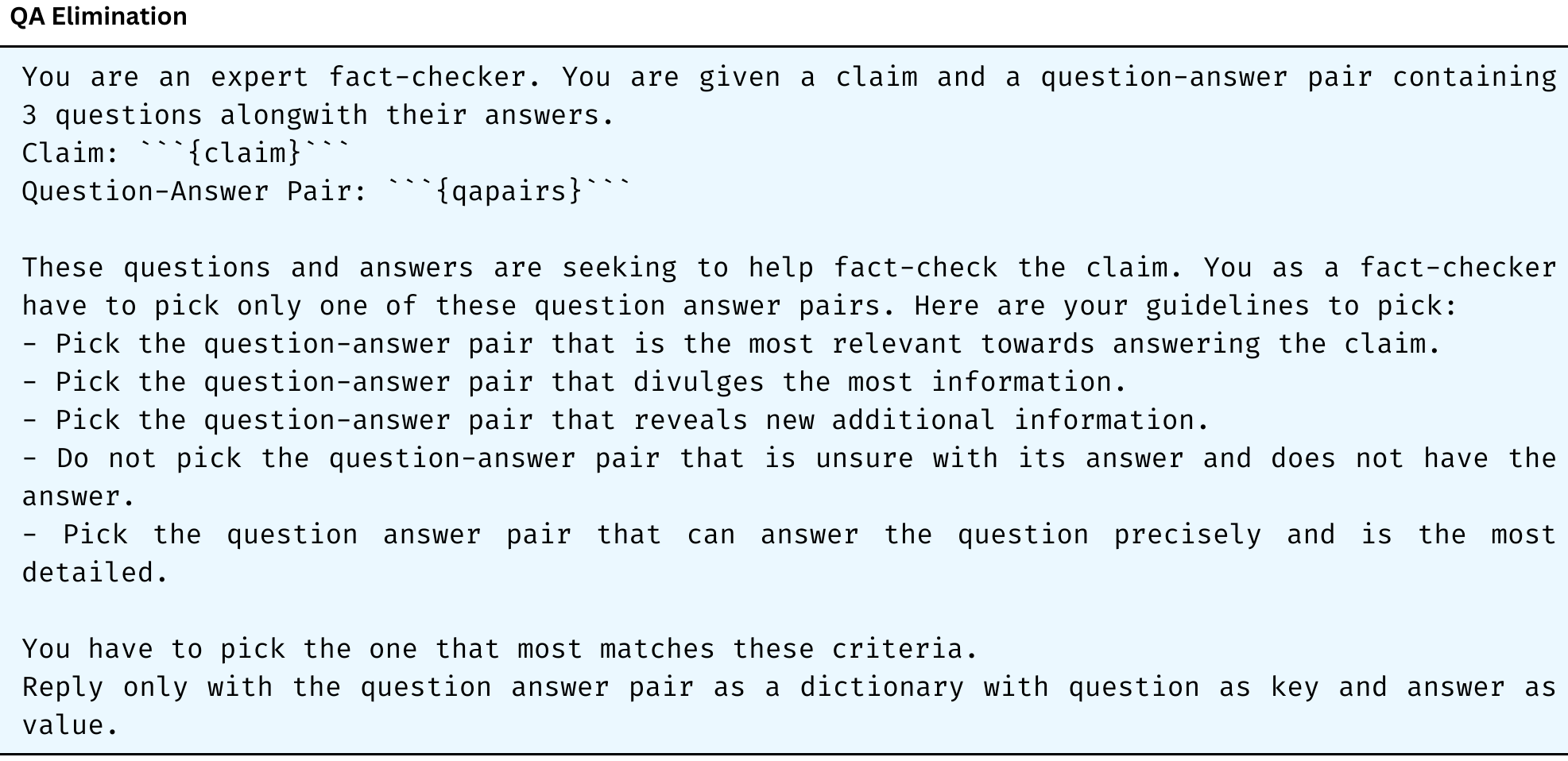}
    \caption{Prompt for QA Elimination}
    \label{fig:qaelim}
\end{figure}

\subsection{Prompts for Veracity Prediction}
\label{prompts_veracity}

\subsubsection{Standard Veracity Prediction Prompt}
\begin{figure}[ht]
    \centering
    \includegraphics[width=14cm]{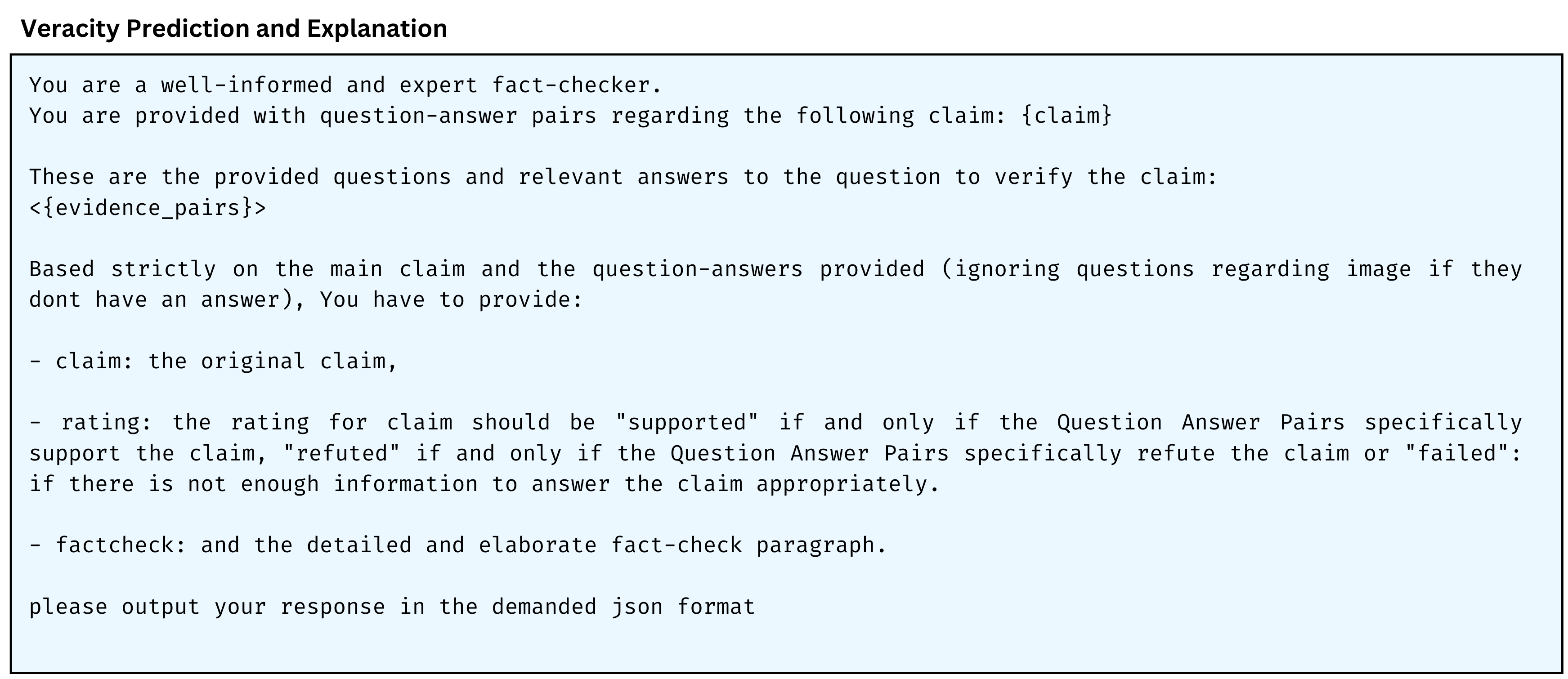}
    \caption{Prompt for Standard Veracity prediction}
    \label{fig:standardveracity}
\end{figure}

\clearpage

\subsubsection{Zero Shot Chain of Thought Veracity Prediction}
\begin{figure}[ht]
    \centering
    \includegraphics[width=14cm]{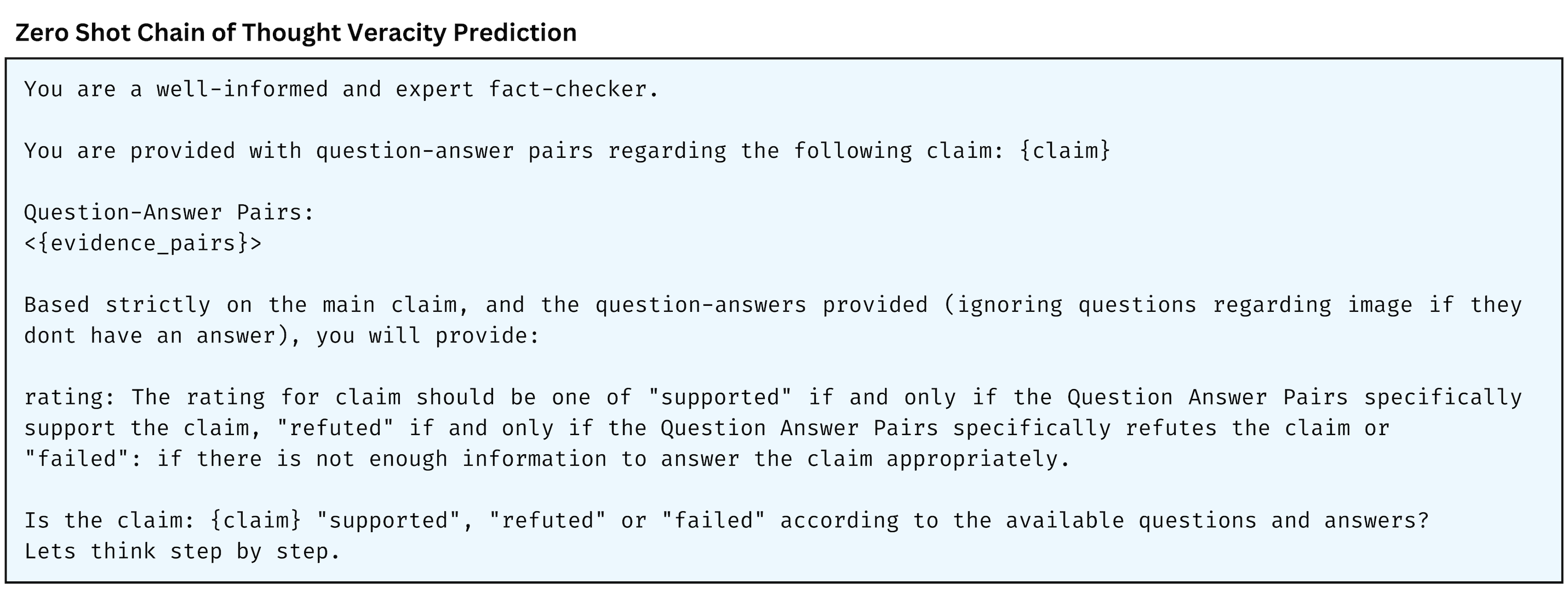}
    \caption{Prompt to get the CoT Veracity Prediction from the question-answer pairs and the claim}
    \label{fig:ZSCOT}
\end{figure}

\subsubsection{Chain of Verification Veracity Prediction}
\label{covesection}

\begin{figure}[ht]
    \centering
    \includegraphics[width=10cm]{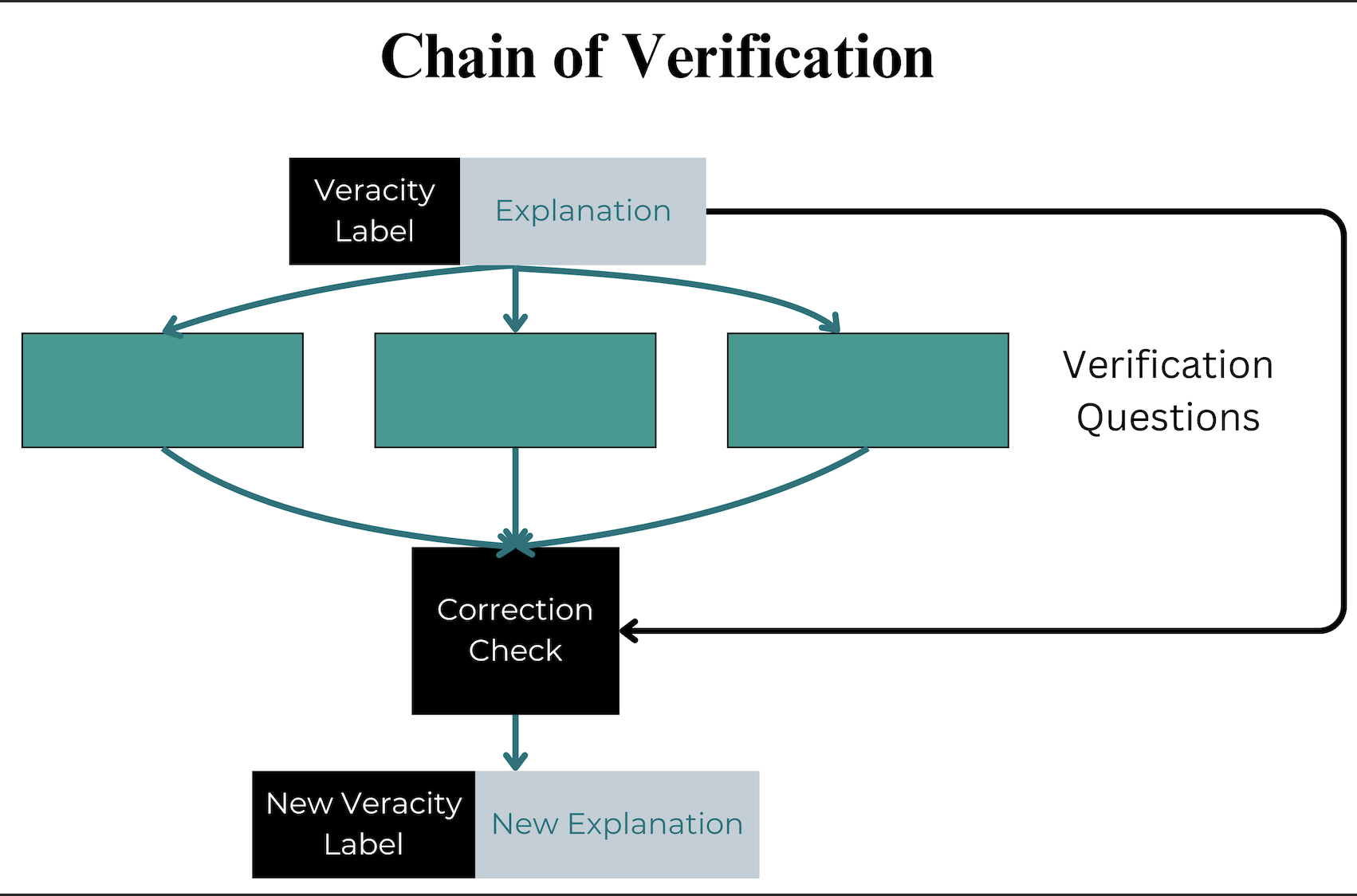}
    \caption{Pipeline of the CoVe Veracity Prediction}
    \label{fig:Pipeline of the CoVe Veracity Prediction}
\end{figure}

\begin{figure}[ht]
    \centering
    \includegraphics[width=14cm]{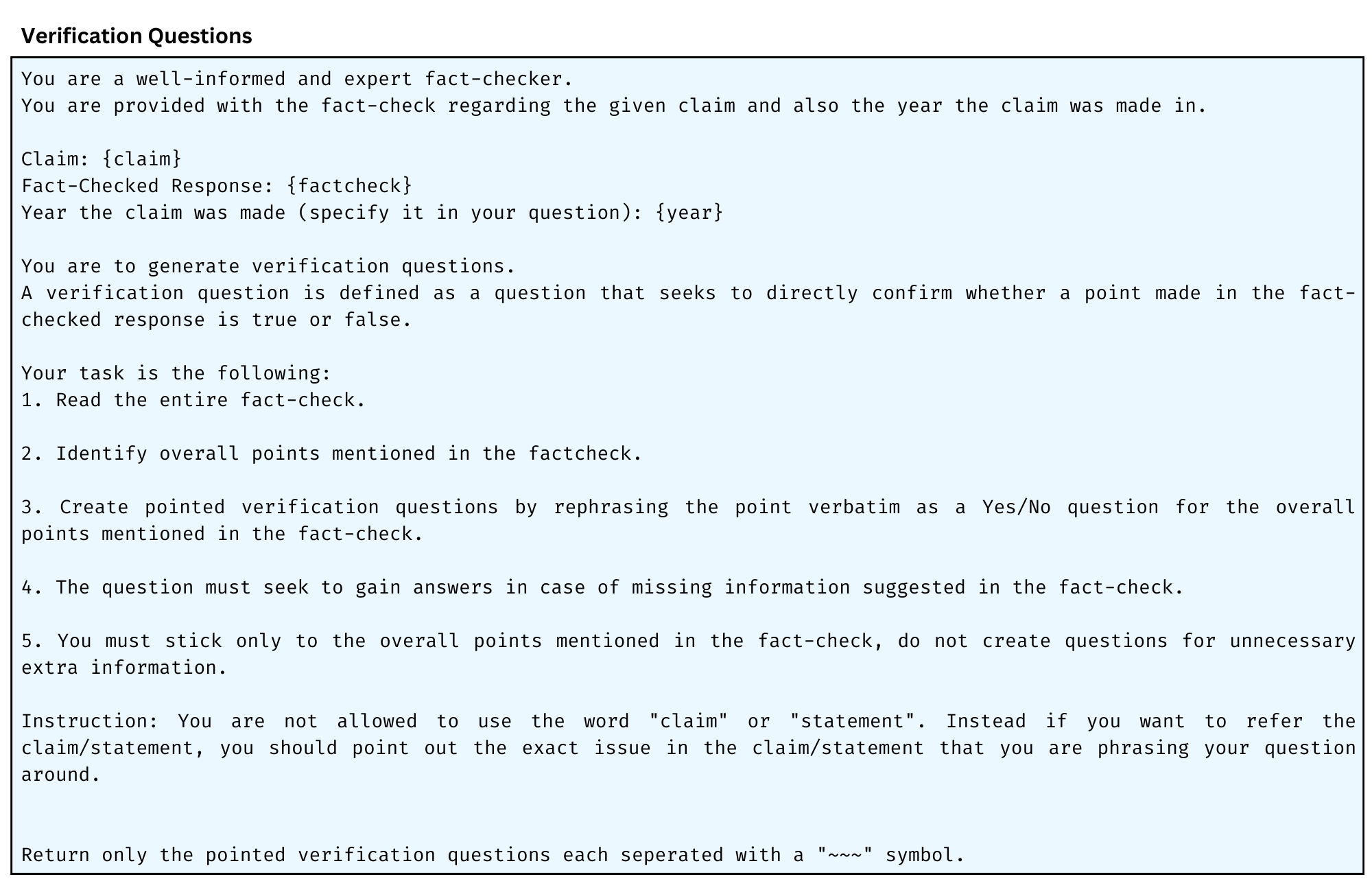}
    \caption{CoVe Verification Questions prompt}
    \label{fig:CoVe Verification Questions prompt}
\end{figure}


\begin{figure}[ht]
    \centering
    \includegraphics[width=14cm]{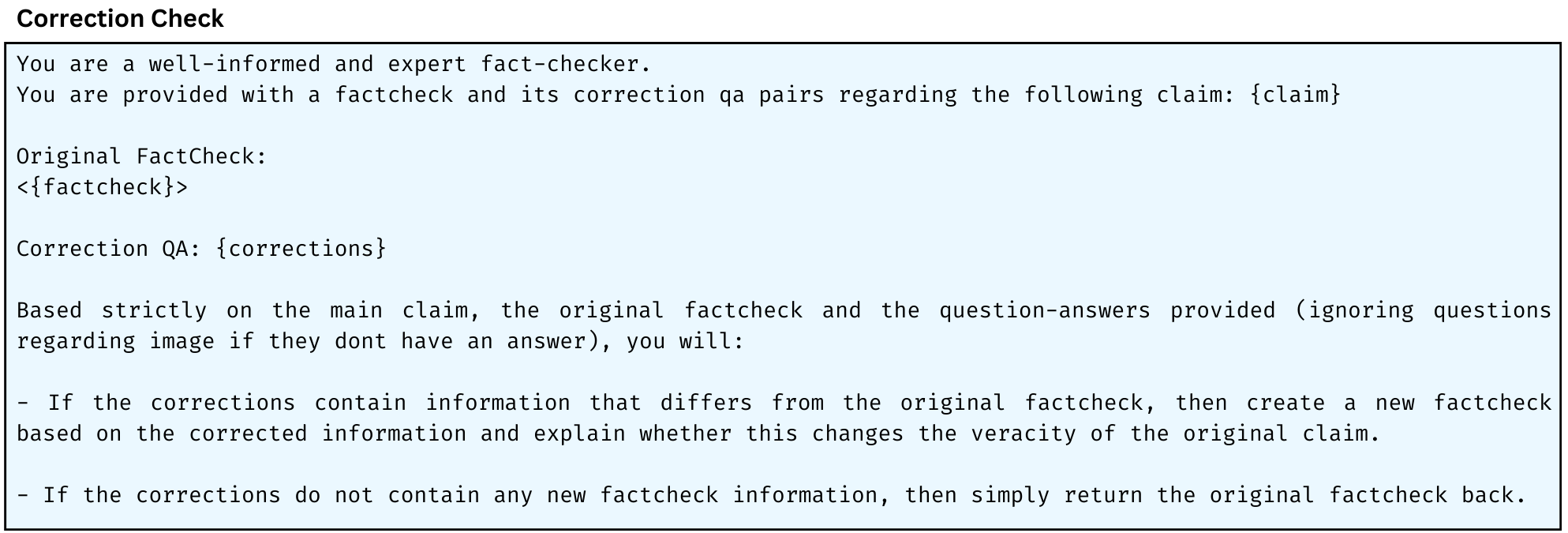}
    \caption{CoVe Corrections Prompt}
    \label{fig:CoVe Corrections Prompt}
\end{figure}

\clearpage

\end{document}